\documentclass[10pt,twocolumn]{article}
\usepackage[letterpaper,top=0.66in,bottom=0.72in,left=0.66in,right=0.66in,
            columnsep=0.25in,headsep=0.18in]{geometry}
\usepackage{newtxtext,newtxmath}
\usepackage{cite}

\usepackage{amsmath,amssymb,amsfonts}
\usepackage{algorithm}
\usepackage{algorithmic}
\usepackage{graphicx}
\usepackage{booktabs}
\usepackage{tabularx}
\usepackage{textcomp}
\usepackage{url}
\usepackage{xcolor}
\usepackage{microtype}
\usepackage{enumitem}
\usepackage{authblk}
\usepackage{titlesec}
\usepackage{caption}
\usepackage{fancyhdr}
\usepackage{placeins}
\usepackage[hidelinks]{hyperref}
\graphicspath{{figures/}}

\definecolor{PreprintBlue}{HTML}{194F7A}
\definecolor{PreprintGray}{HTML}{5E6872}
\hypersetup{
  colorlinks=true,
  linkcolor=PreprintBlue,
  citecolor=PreprintBlue,
  urlcolor=PreprintBlue,
  pdftitle={Numbers Beat Words: A Rigorous On-Premise Benchmark for Coupled MIMO Controller Tuning},
  pdfauthor={Yang Shu, Jiaxuan Chen, Haonan Li},
  pdfkeywords={large language models, controller tuning, PID, multivariable control, process control, benchmarking}
}
\titleformat{\section}
  {\large\bfseries\color{PreprintBlue}}{\thesection}{0.55em}{}
\titleformat{\subsection}
  {\normalsize\bfseries\color{PreprintBlue}}{\thesubsection}{0.5em}{}
\titleformat{\subsubsection}
  {\normalsize\itshape\color{PreprintGray}}{\thesubsubsection}{0.45em}{}
\titlespacing*{\section}{0pt}{1.2ex plus 0.3ex minus 0.2ex}{0.55ex}
\titlespacing*{\subsection}{0pt}{1.0ex plus 0.2ex minus 0.2ex}{0.4ex}
\titlespacing*{\subsubsection}{0pt}{0.8ex plus 0.2ex minus 0.1ex}{0.3ex}

\setlist[itemize]{leftmargin=1.25em,itemsep=0.2ex,topsep=0.4ex}
\setlist[enumerate]{leftmargin=1.4em,itemsep=0.2ex,topsep=0.4ex}

\newcommand{\IEEEPARstart}[2]{#1#2}
\newcommand{\IEEEsetlabelwidth}[1]{}
\newenvironment{IEEEdescription}[1][]
  {\begin{description}[leftmargin=5.2em,labelwidth=4.7em,labelsep=0.5em,
                      style=multiline,font=\normalfont]}
  {\end{description}}
\newenvironment{IEEEkeywords}
  {\par\small\noindent\textbf{\color{PreprintBlue}Keywords---}}
  {\par\normalsize}

\begin{document}
\title{Numbers Beat Words: A Rigorous On-Premise Benchmark for Coupled MIMO Controller Tuning}

\author[2]{Yang Shu\thanks{Yang Shu and Jiaxuan Chen contributed equally to this work.
  Corresponding author: Yang Shu
  (\href{mailto:shuyang@zju.edu.cn}{shuyang@zju.edu.cn}).}}
\author[1]{Jiaxuan Chen}
\author[3]{Haonan Li}
\affil[1]{Jinling Institute of Technology, Nanjing, China}
\affil[2]{Zhejiang University, Hangzhou, China}
\affil[3]{College of Water Resources and Civil Engineering,
  China Agricultural University, Beijing, China}
\date{\small Preprint --- July 2026\\[0.55em]
\begin{minipage}{0.92\textwidth}
\centering\footnotesize\itshape\color{PreprintGray}
This work has been submitted to the IEEE for possible publication.
Copyright may be transferred without notice, after which this version
may no longer be accessible.
\end{minipage}}

\maketitle

\begin{abstract}
Tuning controllers for strongly coupled multi-input multi-output (MIMO) industrial
processes is hard: decentralized classical auto-tuning ignores loop interaction, and
\emph{local} numerical optimization from natural initializations stalls in the resulting
non-convex cost landscape. We ask whether on-premise open-weight large language models
(LLMs) can help, and we built this benchmark to find out honestly---including testing the
classical alternatives that could make the LLM unnecessary, not only the ones that make it
look good. On a single-loop CSTR, classical relay-feedback tuning (IAE~$0.106$) beats an LLM
tuner ($0.162$). On a strongly coupled quadruple-tank with conflicting set-points, scored by
a penalized cost $J=\mathrm{IAE}+\lambda\,\mathrm{TV}(u)$, naive relay and naive LLM tuning
are both no better than open loop, and a local optimizer from balanced starts fails in
$10/10$ runs. A scaffolded LLM given a coupling summary instead reliably ($10/10$) converges
on a counter-intuitive asymmetric structure and, refined by a local optimizer, reaches
$J~12.0\pm0.16$---though a later ablation (Sec.~\ref{sec:ablation}) shows this specific
reliability owes more to an answer-shaped example in the prompt's output format than to
reasoning over the coupling data alone. We then test the most direct classical alternative:
Virtual Reference Feedback Tuning (VRFT), which fits a controller from a single open-loop
experiment, no LLM and no iterative closed-loop search required. Refined the same way, VRFT
matches the hybrid's success rate ($10/10$) with roughly $3\times$ tighter spread and reaches
a \emph{better}
solution ($J~11.1\pm0.05$) at a comparable or lower reported evaluation cost. A Bayesian-optimization
baseline, tested as the standard sample-efficient alternative, likewise does not beat
differential evolution here, and neither black-box method beats VRFT. VRFT's own remaining
weakness is choosing its one design parameter $\tau$: getting it badly wrong costs dearly
before refinement, though we show this shrinks to a narrow low-$\tau$ failure region once
refinement is applied. Reasoning over the same open-loop data VRFT already collects, the LLM
proposes a usable $\tau$ in $8/10$ seeds---but a one-line deterministic rule (the median of the
same rise-time numbers) does the same job in $10/10$ seeds at no extra cost, matching the
LLM's quality; we credit the arithmetic, not the LLM, with removing VRFT's last manual step. On
a benign coupled plant neither the LLM nor VRFT's advantage over naive tuning is needed, and we
show this boundary is predictable in advance: across four structurally different plants, the
relative gain array---computable from step-test data alone, before running any optimization---
tracks, monotonically, whether a structural prior of any kind is worth the investment; a
second, confirmatory signal (the start-sensitivity of a black-box optimizer) requires actually
running a few cheap optimizer attempts but agrees with it on every plant we tested. We
contribute a rigorous benchmark for when on-premise LLMs are worth using for controller
tuning, together with an a priori RGA-based diagnostic, and show that on our central
pathological case a decades-old direct data-driven method is the better default; even its
remaining design choice is better fixed with arithmetic than with an LLM.
\end{abstract}


\section*{Nomenclature}
\begin{IEEEdescription}[\IEEEsetlabelwidth{$\mathrm{TV}(u)$}]
\item[$h_1,h_2$] Controlled lower-tank levels (m).
\item[$v_1,v_2$] Pump command voltages (V), $v_i\in[0.1,10]$.
\item[$G$] Steady-state input--output gain matrix.
\item[$\lambda_{11}$] $(1,1)$ element of the relative gain array (RGA).
\item[$K_{p,i},K_{i,i}$] Proportional/integral gains of loop $i$.
\item[$\mathrm{IAE}$] Integral of absolute tracking error.
\item[$\mathrm{TV}(u)$] Total variation of the control signal (effort).
\item[$\lambda$] Control-effort penalty weight ($=0.75$).
\item[$J$] Penalized cost, $J=\mathrm{IAE}+\lambda\,\mathrm{TV}(u)$.
\end{IEEEdescription}

% ============================================================
\section{Introduction}
\IEEEPARstart{I}{ndustrial} processes are regulated by feedback controllers---most often
PID loops---whose gains must be \emph{tuned}. Tuning is routine for a single loop but
becomes difficult for multivariable plants in which manipulated variables affect several
outputs simultaneously: the loops interact, and tuning each loop in isolation can be
worse than not controlling at all. Classical remedies (decoupling design, model
predictive control) require an accurate process model that is costly to obtain, while
data-driven optimization requires the practitioner to formulate the objective, bounds and
controller structure correctly.

Large language models (LLMs) have recently been applied to control design, typically using
frontier, proprietary models accessed through a cloud API. Two practical obstacles limit
this route in industry: (i) sending live process data to an external API is often
prohibited for security and compliance reasons, and (ii) proprietary endpoints cannot be
audited or deployed offline. On-premise, \emph{open-weight} models avoid both problems,
but it is unclear whether they are competent enough---and, more fundamentally, whether an
LLM contributes anything that a plain numerical optimizer does not.

We therefore organize the paper around four questions. \emph{(Q1) Competence:} can an
on-premise open-weight model tune a controller at all, and how does it compare to a textbook
classical method on a well-posed single loop? \emph{(Q2) Distinctiveness:} on a strongly
coupled multivariable plant, where naive decentralized tuning fails, does the LLM beat a
black-box numerical optimizer---and if so, \emph{why}, mechanistically? \emph{(Q3)
Necessity:} is the LLM actually the best available tool, or does a classical
\emph{direct data-driven} method---designed for exactly this ``no first-principles model''
setting---do as well or better? \emph{(Q4) Boundary:} when, if ever, is the LLM worth the
trouble? Answering these requires resisting the temptation to report only the favorable
case, and Q3 in particular requires being willing to let the LLM lose to a fair alternative.
We show that the answer to Q1 is ``yes, but a good classical tuner is better,'' that Q2's
answer is genuinely positive---the same open model that loses on an easy loop succeeds on a
hard one where naive methods fail, by supplying a \emph{structural prior} (which loop should
dominate) that local optimization cannot recover from naive starts---but that Q3's answer is
sobering: a decades-old direct data-driven method, Virtual Reference Feedback Tuning (VRFT),
beats the LLM's own hybrid pipeline on this same hard case. Q4 then delimits what, if
anything, is left for the LLM once a fair classical alternative is on the table.

\begin{figure*}[t]\centering\includegraphics[width=0.96\textwidth]{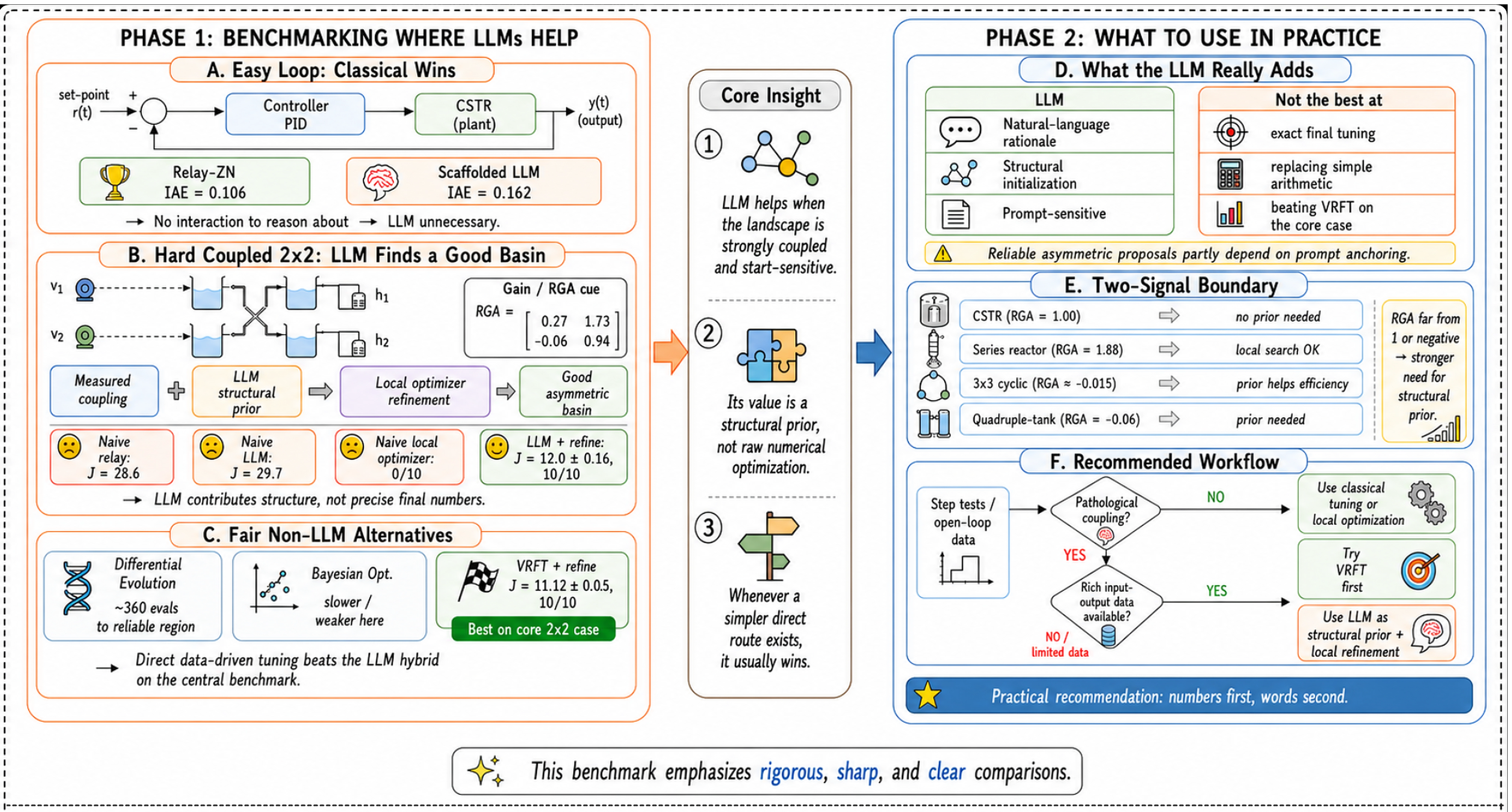}
\caption{Story of the benchmark. (a) On a benign single loop, classical relay tuning is
already near-optimal and the on-premise LLM adds no measurable advantage. (b) On the
pathological coupled quadruple-tank, measured coupling is passed to an on-premise LLM as a
structural scaffold; a numerical optimizer then supplies gain magnitudes. The scaffold can
move a local search from naive spurious minima into a usable basin, but the direct data-driven
VRFT route reaches a lower cost on this central case. (c) The boundary is screened with RGA
and inexpensive optimizer start-sensitivity. The $3\times3$ efficiency statement is an
LLM-versus-differential-evolution comparison only; VRFT was not rerun at $3\times3$.}
\label{fig:concept}\end{figure*}

This paper answers those questions with a benchmark designed to expose failure---the LLM's
own failure to be necessary, included---as readily as success (Fig.~\ref{fig:concept}). We
do not assume the LLM is superior; we measure where it helps, where it does not, and where
something simpler beats it outright. Our findings are:
\begin{itemize}
\item \textbf{On simple loops, classical methods win.} On a CSTR, relay-feedback
Ziegler--Nichols tuning (IAE~$0.106$) is near-optimal ($0.102$) and beats a scaffolded LLM
tuner ($0.162$): for well-posed single loops the LLM is unnecessary.
\item \textbf{On strongly coupled MIMO, all naive methods fail.} On the quadruple-tank with
conflicting set-points, scored by the effort-penalized cost $J=\mathrm{IAE}+\lambda\,\mathrm{TV}(u)$,
decentralized relay tuning ($J~28.6$) and naive LLM tuning ($J~29.7$) are both no better
than open loop ($J~22.7$), and a black-box optimizer is \emph{unreliable}, failing to reach
the reference region in all $10$ runs started from natural balanced gains (mean $J~28.0$).
\item \textbf{The scaffolded LLM beats naive baselines, but its specific structure is prompt-sensitive.}
A scaffolded open LLM reasons about the measured coupling and proposes the counter-intuitive
asymmetric structure, reaching $J~16.9\pm0.2$ consistently across ten decoding seeds, and the
coordinated-region benefit holds across four open models. A follow-up ablation
(Sec.~\ref{sec:ablation}), prompted by reviewer scrutiny, found that our original ablation
undercontrolled for the prompt's required-JSON format example, which itself encodes a specific
asymmetric direction; with that example also neutralized, the pipeline still clearly beats
naive baselines but no longer reliably converges to the one specific structure we report
throughout. We correct the claim accordingly rather than remove the finding.
\item \textbf{The hybrid is reliable but not best on this plant---and a classical method beats it.} The
LLM fixes a structural basin (which loop dominates) that \emph{local} optimization cannot
find from naive starts; the optimizer then refines magnitudes within that basin---including a
non-obvious \emph{negative} integral correction. The hybrid reliably ($10/10$) reaches
$J~12.0\pm0.16$. But Virtual Reference Feedback Tuning (VRFT)---fit from a single open-loop
experiment, no LLM, no closed-loop search---refined the same way matches the hybrid's success
rate ($10/10$) with roughly $3\times$ tighter spread and reaches a \emph{better} solution
($J~11.1\pm0.05$).
We did not have to look hard for this: it is what a control engineer would try first.
\item \textbf{Neither black-box method dominates VRFT, and the LLM's real edge is narrower
than efficiency.} A global optimizer (differential evolution) and a standard
Bayesian-optimization baseline both reach the coupled-plant reference region no more cheaply
than VRFT, and BO does not even beat DE here. What the LLM retains is that it needs no
per-plant reference-model hyperparameter in its scaffolding prompt, and it returns a stated
rationale alongside its proposal; VRFT's own quality swings from $J~14.2$ to $J~29.7$ across
its $\tau$ scan \emph{before} refinement, though we show below this shrinks substantially once
refinement is applied. Against the global optimizer specifically---the comparison we could
extend to a higher-dimensional plant---this efficiency gap \emph{widens} with dimension: on a
$3\times3$ plant the hybrid reaches the near-optimal region in $\sim\!8\times$ fewer evaluations than DE.
We did not re-run VRFT at $3\times3$, so we do not know whether this specific margin would
survive; we report it as a property of the LLM-vs-DE comparison, not of the LLM-vs-VRFT one.
\item \textbf{Even VRFT's one weakness is fixed by arithmetic, not the LLM.} $\tau$-sensitivity
survives refinement only at the extreme low end; across a full order of magnitude
($\tau\in[60,600]$s) refinement reaches the same near-optimal basin regardless of the exact
value. Reasoning over the same open-loop data VRFT already collects, the LLM proposes a usable
$\tau$ in $8/10$ seeds ($J=11.19\pm0.17$ on the successes); but the median of the same
rise-time numbers, used directly with no reasoning step, succeeds in $10/10$ seeds at
$J=11.17\pm0.15$---matching the LLM's quality, beating its reliability, at zero marginal cost.
We report this as a negative result for the LLM by design: every time we gave it a fair,
simpler competitor, something simpler won.
\item \textbf{Whether a structural prior helps at all is predictable in advance.} Across four
structurally different plants---a trivial single loop, our pathological quadruple-tank, a
benign interacting series reactor, and a $3\times3$ cyclic plant deliberately added to probe
the space between---the RGA diagonal and the start-sensitivity of a naive local optimizer move
together, monotonically: the further the RGA is from ideal pairing, the more naive-start local
search fails, and the more a structural prior (LLM or VRFT) is needed. RGA is genuinely free,
computable from step-test data before running anything; start-sensitivity requires a few
cheap optimizer attempts but agrees with RGA on every plant we tested, so a practitioner can
decide, from RGA alone or with this inexpensive confirmatory check, whether the investment is
worth making on their own plant.
\end{itemize}
All code, data and prompts are released for reproducibility.

% ============================================================
\section{Related Work}
\textbf{LLMs for control design.} A fast-growing line of work casts large language models as
agents that synthesize or tune controllers. ControlAgent~\cite{controlagent} integrates a
central coordinator with task-specific LLM agents and encodes control-oriented domain
expertise, emulating an engineer's iterative loop to meet stability, performance and
robustness specifications; AgenticControl~\cite{agenticcontrol} similarly automates control
design through a multi-agent LLM framework. Peer-reviewed work has begun to test related
ideas in narrower settings: SmartControl~\cite{smartcontrol} combines LLM-mediated
requirement interpretation and response evaluation with particle-swarm and differential-
evolution gain search for interactive PID design, while Wen \emph{et al.}~\cite{wen2026tank}
experimentally study an LLM in a water-tank control loop. The two agentic frameworks above
are compelling but share three
features that our study deliberately departs from. First, they rely on \emph{proprietary
frontier} models behind a cloud API, which is precisely what process-industry data-governance
rules forbid and what cannot be audited or deployed offline. Second, they are evaluated
mostly on \emph{linear, single-loop or frequency-domain} design criteria, leaving open how
LLMs behave on strongly coupled, nonlinear MIMO plants where decentralized intuition breaks
down. Third---and most importantly for the scientific question---they do not isolate
\emph{whether the LLM contributes anything a plain numerical optimizer does not}; an LLM that
merely searches the gain space could be replaced by Nelder--Mead. We address all three: we
benchmark \emph{on-premise open} models, on a pathological coupled MIMO plant, and we pit the
LLM head-to-head against a black-box optimizer to identify its distinctive contribution as a
\emph{structural prior} rather than a searcher. A parallel line treats LLMs as black-box
optimizers that propose-and-evaluate candidate solutions in natural language
(OPRO~\cite{opro}); we instead use the LLM not to optimize numerically but, leveraging its
step-by-step reasoning ability~\cite{cot}, to supply a \emph{structural} prior that a
numerical optimizer then refines. We target precisely these
gaps, on the open-source PC-Gym process-control benchmark~\cite{pcgym}.

\textbf{Classical and optimization-based tuning.} Relay-feedback auto-tuning
(\AA str\"om--H\"agglund)~\cite{astrom1984} and Ziegler--Nichols rules~\cite{zn1942} remain
the workhorses for single loops because they identify the critical point directly and need
no model (see~\cite{borase2021,somefun2021} for recent surveys of the field); analytic
model-based rules such as SIMC~\cite{simc} refine this with an explicit
performance/robustness trade-off, on the classical feedback foundations of~\cite{astrommurray}.
For multivariable plants the relative gain array~\cite{bristol1966} quantifies
loop interaction and guides input--output pairing, while decoupling design and model
predictive control~\cite{mpcbook} explicitly handle interaction but require an accurate
process model~\cite{skogestad2005} that is often expensive to obtain and maintain. The
quadruple-tank process~\cite{johansson2000} is a canonical multivariable benchmark whose
adjustable transmission zero can place the plant in a non-minimum-phase, hard-to-decouple
regime; we operate it with conflicting set-points precisely to make decentralized tuning
fail. When no model is available, practitioners fall back on direct numerical optimization
of the controller gains against a simulated or measured cost, from derivative-free global
search such as differential evolution~\cite{stornprice} to Bayesian optimization with safety
constraints~\cite{konig2022,duivenvoorden2017}. A separate, longer-standing branch of
control avoids both first-principles modeling and iterative closed-loop search altogether:
\emph{direct data-driven controller tuning}, which fits controller parameters from a single
batch of measured input--output data using a reference-model matching criterion. Iterative
Feedback Tuning~\cite{hjalmarsson1998} refines this iteratively from closed-loop
experiments, while Virtual Reference Feedback Tuning (VRFT)~\cite{campi2002vrft} obtains a
controller from one open-loop (or arbitrary) experiment via a single linear least-squares
fit; the original formulation is single-loop, and we apply it decentrally, one fit per paired
loop (Sec.~\ref{sec:vrft})---we do not implement or claim to test a fully multivariable VRFT
extension. This family is the most direct classical
competitor to an LLM-based structural prior, since it shares the same ``no first-principles
model'' premise; we benchmark VRFT against our LLM pipeline directly in
Sec.~\ref{sec:vrft}, rather than only against numerical search. Local numerical optimization
of the kind above is powerful and model-free, but---as our reliability experiment shows---the
cost landscape of a strongly coupled plant is non-convex and riddled with spurious minima, so
the outcome depends heavily on initialization. Our work does not replace these tools; it adds a component (a reasoned
structural prior) that makes the \emph{optimization} step dependable where it otherwise is
not.

\textbf{On-premise open LLMs.} The rapid maturation of open-weight models such as
Qwen3~\cite{qwen3}, Qwen2.5~\cite{qwen25} and Llama~3~\cite{llama3} makes offline, auditable
deployment practical: a 14B-parameter model runs on a single workstation GPU and keeps all
process data on site. This is a hard prerequisite for industrial control, where streaming
live plant measurements to a third-party API is frequently disallowed, yet it is exactly the
setting that proprietary-API control studies sidestep. A natural worry is that open models
are too weak to be useful here; part of our contribution is to show that even where their
raw numerical tuning is unremarkable, their \emph{reasoning} about plant structure is strong
and consistent enough to be the decisive ingredient---and that this holds across model
families and sizes, not just for one flagship checkpoint.

% ============================================================
\section{Problem Setup and Methods}
\subsection{Benchmarks}
We use the PC-Gym process-control simulators~\cite{pcgym}. The \emph{CSTR} is a single-loop
set-point tracking task (control concentration $C_a$ via coolant temperature). The
\emph{quadruple-tank}~\cite{johansson2000} is a $2\times2$ strongly coupled plant. Its four
liquid levels evolve according to the mass-balance dynamics
\begin{align}
\dot h_1 &= -\tfrac{a_1}{A_1}\sqrt{2gh_1}+\tfrac{a_3}{A_1}\sqrt{2gh_3}+\tfrac{\gamma_1 k_1}{A_1}v_1, \nonumber\\
\dot h_2 &= -\tfrac{a_2}{A_2}\sqrt{2gh_2}+\tfrac{a_4}{A_2}\sqrt{2gh_4}+\tfrac{\gamma_2 k_2}{A_2}v_2, \nonumber\\
\dot h_3 &= -\tfrac{a_3}{A_3}\sqrt{2gh_3}+\tfrac{(1-\gamma_2)k_2}{A_3}v_2, \label{eq:fourtank}\\
\dot h_4 &= -\tfrac{a_4}{A_4}\sqrt{2gh_4}+\tfrac{(1-\gamma_1)k_1}{A_4}v_1, \nonumber
\end{align}
where $A_i,a_i$ are tank and outlet cross-sections, $k_i$ the pump constants and
$\gamma_i\in(0,1)$ the valve split ratios; pump $v_1$ feeds tanks $1$ and $4$, pump $v_2$
feeds tanks $2$ and $3$. The square-root outflows make the plant \emph{nonlinear}, and the
upper tanks $3,4$ couple each pump into the \emph{opposite} lower level. We control the
lower tanks $h_1,h_2$ with $v_1,v_2$. Numerically identifying the steady-state gain matrix about
the operating point (perturbing each pump and measuring both levels) gives
\begin{equation}
G=\begin{bmatrix} \partial h_1/\partial v_1 & \partial h_1/\partial v_2 \\ \partial h_2/\partial v_1 & \partial h_2/\partial v_2 \end{bmatrix}
=\begin{bmatrix} 0.005 & 0.024 \\ 0.029 & 0.008 \end{bmatrix},
\end{equation}
whose off-diagonal entries dominate ($v_1$ chiefly affects $h_2$, $v_2$ chiefly affects
$h_1$). The corresponding relative gain array~\cite{bristol1966} has $\lambda_{11}=-0.06$:
the \emph{negative} diagonal RGA element signals that any decentralized controller assuming
the natural $v_1\!-\!h_1$, $v_2\!-\!h_2$ pairing faces a closed-loop integrity problem,
quantifying why per-loop tuning fails here. We further impose \emph{conflicting} set-points
($h_1$ up while $h_2$ down) so the two loops fight through the coupling. Controllers are
position-form PI loops, $u(t)=u_{\mathrm{bias}}+K_p\,e(t)+K_i\!\int_0^t\!e(\tau)\,d\tau$ per
loop with the cross pairing, with clamping anti-windup (the integral term is held, not
advanced, whenever the unclipped output would exceed the actuator bound). To reward controllers
that track \emph{without} excessive actuation, performance is a penalized cost
$J=\mathrm{IAE}+\lambda\,\mathrm{TV}(u)$, where $\mathrm{IAE}=\mathrm{IAE}_{h_1}+\mathrm{IAE}_{h_2}$
is the total tracking error and $\mathrm{TV}(u)=\sum_t|\Delta v_1|+|\Delta v_2|$ (scaled by
the actuator span) is the total variation of the two pump signals, a standard proxy for
control effort and valve wear. We set $\lambda=0.75$: this is the smallest weight at which
the optimal controller leaves the aggressive, near-saturating ``bang--bang'' regime and
enters a smooth-actuation regime, so the penalty acts as a mild
regularizer (it contributes under $10\%$ of $J$ at the optimum) rather than dominating the
objective. Reporting $J$ ensures the tuners are not rewarded for chattering actuators.

\subsection{Baselines}
(i) \emph{Decentralized relay-feedback Ziegler--Nichols}~\cite{astrom1984,zn1942}: each loop
is tuned independently by a relay experiment about its operating point, which induces a
sustained oscillation whose amplitude and period yield the ultimate gain and period and
hence the ZN PI settings. This is the standard decentralized auto-tuner and ignores the
cross-coupling by construction. (ii) \emph{Local optimizer}: Nelder--Mead on all four
gains jointly, started from a grid of ten naive ``balanced'' gain vectors (equal emphasis
on both loops); we report the distribution over starts to expose start-sensitivity.
(iii) \emph{Global optimizers}: to test whether the optimizer's failure is merely an
artifact of local search, we add a global, derivative-free baseline---differential
evolution over the full gain bounds---and a ten-start Latin-hypercube Nelder--Mead that
explicitly samples asymmetric initializations, both under budgets comparable to the
LLM-plus-refinement pipeline. (iv) \emph{Naive LLM}: an open LLM proposes gains with
iterative $J$ feedback but no process knowledge in the prompt. (v) \emph{Bayesian
optimization}: a Gaussian-process, expected-improvement black-box optimizer
(\texttt{scikit-optimize}), the standard sample-efficient alternative to differential
evolution, run under the same gain bounds. (vi) \emph{Virtual Reference Feedback Tuning
(VRFT)}~\cite{campi2002vrft}: the most direct classical alternative to an LLM-based
structural prior, since it too requires no first-principles model---only a single batch of
input--output data. Sec.~\ref{sec:vrft} describes it and reports its results in full.

\subsection{Scaffolded LLM tuner}
The LLM is given (a) the measured input--output coupling from step tests, (b) per-round
feedback of the achieved $J$ with its IAE/TV breakdown for recent proposals, (c) a generic
engineering heuristic that balanced tuning tends to fail on this plant and that one loop
typically should dominate (without saying \emph{which} loop or by how much), and (d) an
instruction to explore boldly (diverse, asymmetric proposals) while keeping actuation
smooth. It is \emph{not} told the solution. Each round it returns four gains as a JSON
object; we simulate the closed loop and feed the result back, for 18 rounds at sampling
temperature $0.9$. The complete, verbatim system prompt is:
\begin{quote}\small\itshape
``You tune a $2\times2$ MIMO controller for the strongly-coupled quadruple-tank. Two PI
loops: $v_1$ controls $h_2$, $v_2$ controls $h_1$. Pumps $v$ in $[0.1,10]$, gains positive.
Measured coupling: increasing $v_1$ raises $h_2$ strongly and $h_1$ weakly; increasing $v_2$
raises $h_1$ strongly and $h_2$ moderately. Set-points conflict: $h_1$ up, $h_2$ down.
Objective $J=$ IAE (tracking error) $+$ control effort (total pump movement). So you must
track well without aggressive/oscillatory/saturating control---smooth control is rewarded.
Over-large gains cause bang-bang and hurt $J$. Insight: balanced tuning fails; the best
controller emphasizes \emph{one} loop while keeping the other weak. Use moderate gains for
smoothness. Explore boldly but keep gains moderate (avoid huge values that saturate). Reply
short reasoning then JSON last line:
\texttt{\{"Kp1":200,"Ki1":10,"Kp2":8,"Ki2":0.5\}}.''
\end{quote}
We additionally test an ablation (Sec.~\ref{sec:ablation}) that removes even the generic
``emphasize one loop'' phrasing, leaving only the measured coupling facts, to check whether
the structural insight is inferred or merely echoed.

\subsection{Hybrid: LLM structural prior $+$ optimizer refinement}
\label{subsec:hybridmethod}
The LLM's proposed gains define a \emph{structural prior}---which loop to emphasize and the
rough magnitude ratio---rather than a finished controller. We initialize Nelder--Mead from
each of the LLM's ten converged proposals and let it refine all four gains locally. This
separates the two hard sub-problems: identifying the correct (counter-intuitive) basin,
which the optimizer cannot do from naive starts, and polishing magnitudes within it, which
the LLM does not do precisely. The combination is what reliably attains a smooth, near-optimal
solution.
Algorithm~\ref{alg:tune} summarizes the full scaffolded-plus-hybrid procedure.

\begin{algorithm}[t]
\caption{Scaffolded LLM tuning with optimizer refinement}
\label{alg:tune}
\begin{algorithmic}[1]
\REQUIRE plant simulator, measured coupling summary $C$, rounds $R$, seeds $S$, penalty $\lambda$
\STATE \textbf{Identify coupling:} step-test each input, record gain matrix $G$ and RGA
\FOR{each seed $s \in S$}
  \STATE history $\gets \emptyset$, \; $g^\star_s \gets \textsc{null}$
  \FOR{round $= 1$ to $R$}
    \STATE prompt $\gets$ system($C$, ``emphasize one loop, keep control smooth'') $+$ history
    \STATE $g \gets \textsc{ParseJSON}(\textsc{LLM}(\text{prompt}))$ \quad // four PI gains
    \STATE $J,\mathrm{IAE},\mathrm{TV} \gets \textsc{Simulate}(g,\lambda)$
    \STATE append $(g,J,\mathrm{IAE},\mathrm{TV})$ to history; update best $g^\star_s$
  \ENDFOR
\ENDFOR
\STATE \textbf{Scaffolded LLM output:} $\{g^\star_s\}_{s\in S}$ \quad (coupling-conditioned scaffold)
\FORALL{$g^\star_s$}
  \STATE \textbf{Hybrid refinement:}
    $g^{\mathrm{hyb}}_s \gets \textsc{NelderMead}(J;\,\text{init}=g^\star_s)$
\ENDFOR
\RETURN $\arg\min_s J(g^{\mathrm{hyb}}_s)$
\end{algorithmic}
\end{algorithm}

\subsection{Experimental setup}
All models run \emph{on-premise} on a single workstation; a 14B model fits comfortably on
one NVIDIA RTX~6000 Ada (48~GB) GPU, so no data ever leaves the machine.
Open-weight checkpoints (Qwen3-14B, Qwen2.5-7B/14B, Llama-3.1-8B) are served locally through
the HuggingFace \texttt{transformers} runtime in \texttt{bfloat16}; we decode with
temperature $0.9$ and nucleus sampling $p{=}0.95$, and parse the final JSON object from each
reply. Each tuning run is $18$ rounds; the scaffolded tuner is evaluated over ten random
seeds. The quadruple-tank episode uses $N{=}120$ control steps over a $400$\,s horizon with
conflicting set-point changes at the mid-point; pumps are saturated to $[0.1,10]$\,V. The
relay-feedback baseline performs an Åström--Hägglund relay experiment about the operating
point of each loop to extract the ultimate gain/period, then applies the ZN PI rule. The
black-box optimizer and the optimizer stage of the hybrid both use the Nelder--Mead simplex
method~\cite{neldermead} (\texttt{scipy}~\cite{scipy}) with up to $250$--$300$ iterations.
Reference optima are obtained by multi-start Nelder--Mead. All scripts, prompts and the per-run ledger are released for
reproducibility.

% ============================================================
\section{Experiments}
Unless stated otherwise we use Qwen3-14B as the LLM, served on-premise; all numbers come
from the released experiment ledger. Every ``$\pm$'' figure in this paper is a sample
standard deviation across the stated independent runs (seeds), not a standard error or
confidence interval; where we report a confidence interval instead, we say so explicitly and
use the Wilson score interval (see the discussion below Table~\ref{tab:main} for why we do
not treat every reported interval as a population-level inference). Table~\ref{tab:main}
summarizes the coupled-MIMO results that constitute our central evidence.

\begin{table}[t]
\centering
\caption{Coupled MIMO (quadruple-tank, conflicting set-points), penalized cost
$J=\mathrm{IAE}+\lambda\,\mathrm{TV}(u)$ with $\lambda=0.75$; lower is better. IAE and TV
are the tracking and control-effort components. ``Reliable'' is the fraction of runs
reaching the optimum region ($J<20$).}
\label{tab:main}
\small
\begin{tabularx}{\columnwidth}{@{}>{\raggedright\arraybackslash}Xcccc@{}}
\toprule
\textbf{Method} & \textbf{$J$} & \textbf{IAE} & \textbf{TV} & \textbf{Reliable} \\
\midrule
No control                         & 22.7 & 22.7 & 0.0 & --- \\
Naive decentralized relay-ZN       & 28.6 & 21.7 & 9.3 & fails \\
Naive LLM (no process knowledge)   & 29.7 & 23.9 & 7.8 & fails \\
Black-box optimizer (naive starts) & 28.0$^{\dagger}$ & --- & --- & 0/10 \\
Scaffolded LLM (ours)              & 16.9 & 11.6 & 7.0 & 10/10 \\
\textbf{Hybrid: LLM prior $+$ optimizer (ours)} & \textbf{12.0} & \textbf{9.1} & \textbf{3.9} & \textbf{10/10} \\
\midrule
Asymmetric-basin reference$^{\ddagger}$   & 11.9 & 9.1 & 3.8 & --- \\
\bottomrule
\multicolumn{5}{@{}p{\columnwidth}@{}}{\footnotesize
$^{\dagger}$Mean over 10 naive starts; range $23.4$--$30.2$, none $<20$.
$^{\ddagger}$Best point found in the asymmetric basin our method targets; a
separate, lower-$J$ (better) basin exists and is reliably reached by VRFT,
not by our LLM pipeline (Secs.~\ref{sec:landscape-note}, \ref{sec:vrft}).
$J$ is computed from full-precision IAE and TV; components are rounded.}
\end{tabularx}
\end{table}

\subsection{Simple loop: classical beats LLM (CSTR)}
\label{subsec:cstr}
\begin{figure}[t]\centering\includegraphics[width=0.78\columnwidth]{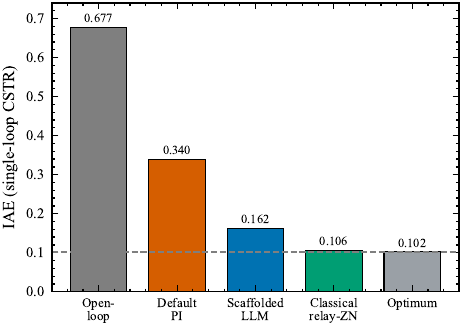}
\caption{CSTR (single loop): classical relay-ZN is near-optimal and beats the LLM.}
\label{fig:cstr}\end{figure}

We begin with Q1 on the single-loop CSTR (Fig.~\ref{fig:cstr}). Relay-feedback
Ziegler--Nichols reaches IAE~$0.106$, essentially the $0.102$ best value found by multi-start
numerical search, whereas the scaffolded LLM, given the same closed-loop feedback over the
same number of rounds, settles at IAE~$0.162$---competent (it tracks the set-point and is
stable) but about $50\%$ worse than the classical tuner and never better across seeds. The
reason is structural: a single loop has no interaction to reason about, so the LLM's
distinctive capability is idle, and a half-century-old relay rule that directly identifies
the critical point is hard to beat. This negative result matters because it disciplines the
rest of the paper: it rules out the trivial explanation that our
open model is simply a strong black-box optimizer dressed up, and it sets up the inversion
on coupled plants, where the very capability that is wasted here becomes valuable.

\subsection{Coupled MIMO: naive methods fail}
\begin{figure}[t]\centering\includegraphics[width=\columnwidth]{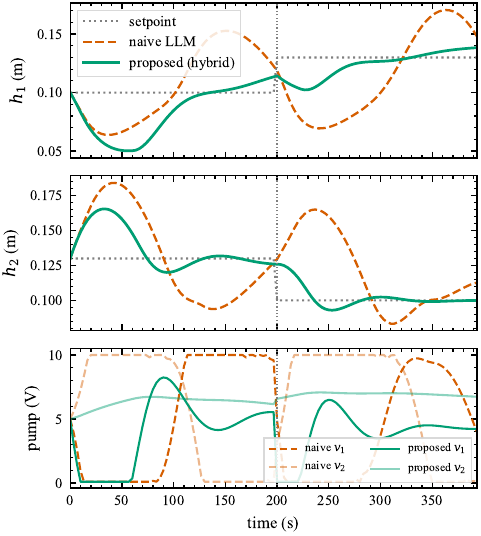}
\caption{Time-domain response (quadruple-tank, conflicting set-points switched at the
dotted line). The naive LLM controller (orange) oscillates and saturates both pumps; the
proposed hybrid controller (green) tracks both levels with smooth, unsaturated control.}
\label{fig:main2}\end{figure}

\begin{figure}[t]\centering\includegraphics[width=0.92\columnwidth]{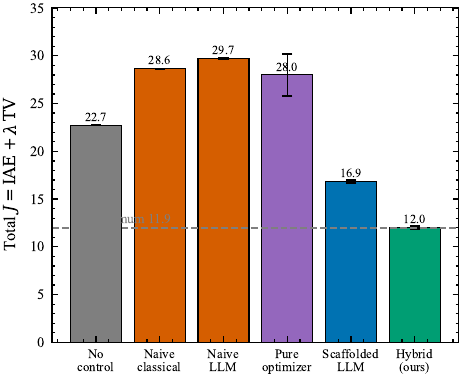}
\caption{Coupled MIMO, penalized cost $J$: naive classical and naive LLM are worse than
open loop and the local optimizer from naive starts is unreliable; the scaffolded LLM is a
large, reliable improvement and the hybrid reaches a near-optimal basin (dashed line)---not
the best one on this plant; Sec.~\ref{sec:vrft} reports a better basin, found reliably by
VRFT instead.}\label{fig:main}\end{figure}

On the quadruple-tank with conflicting set-points, decentralized relay tuning yields
$J~28.6$ (IAE~$21.7$, TV~$9.3$)---worse than open loop ($J~22.7$)---because tuning each
loop in isolation ignores the dominant cross-coupling, so the loops fight and the
actuators thrash. Naive LLM tuning is no better ($J~29.7$): given only cost feedback it
does not explore, locking onto a balanced, textbook guess. Fig.~\ref{fig:main2} shows the
time-domain consequence: the naive controller leaves the set-points poorly tracked while
its pumps saturate (bang--bang), whereas the proposed controller tracks smoothly with
modest actuation. Fig.~\ref{fig:main} places all methods on the penalized cost.

\subsection{Reliability: local optimization is start-dependent, the LLM is not}
\begin{figure}[t]\centering\includegraphics[width=0.92\columnwidth]{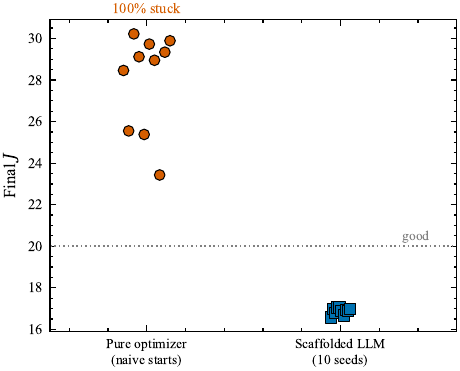}
\caption{Penalized cost $J$ across runs. The plain optimizer from naive starts scatters
high (0/10 reach the good region); the scaffolded LLM clusters tightly and reliably across
ten seeds.}\label{fig:rel}\end{figure}

A \emph{local} optimizer---the default in many tuning toolchains---can in principle reach the
optimum, but on the penalized landscape it does so only from a fortunate initialization.
Started from ten natural ``balanced'' gain vectors, Nelder--Mead fails to reach the optimum
region in \emph{all} $10$ runs (mean $J~28.0$, range $23.4$--$30.2$, none below $20$); the
coordinated solution is counter-intuitive (emphasize one loop, keep the other weak, with a
small negative integral correction) and lies outside the basin of balanced starts. (A
\emph{global} optimizer does reach it; we examine that comparison and the LLM's
sample-efficiency advantage in Sec.~\ref{sec:global}.) The scaffolded LLM, by reasoning
about the measured coupling, proposes exactly that asymmetric structure and reaches
$J~16.9\pm0.2$ over ten decoding seeds, \emph{without depending on a lucky numerical
initialization the way local optimization does} (Fig.~\ref{fig:rel}). This
reliability---the LLM consistently identifying the right structural basin regardless of
sampling seed---is its distinctive contribution; the optimizer then closes the remaining magnitude gap
(Sec.~\ref{sec:hybrid}). It is worth dwelling on why this is the crux of the paper rather
than the raw $J$ values. A practitioner does not get ten tries with an oracle that tells them
which run succeeded; they get one tuning attempt, and they need it to land in a usable
region. Local refinement from a naive start offers no such guarantee---its outcome is a
lottery over the start---whereas the scaffolded LLM returns a good controller on essentially
every attempt, with a seed-to-seed spread ($\pm0.2$ in $J$ over ten seeds) an order of
magnitude smaller. Reliability, not peak performance, is what makes a tuner trustworthy in
deployment, and it is exactly what the structural prior buys without a global search.

\begin{figure}[t]\centering\includegraphics[width=0.96\columnwidth]{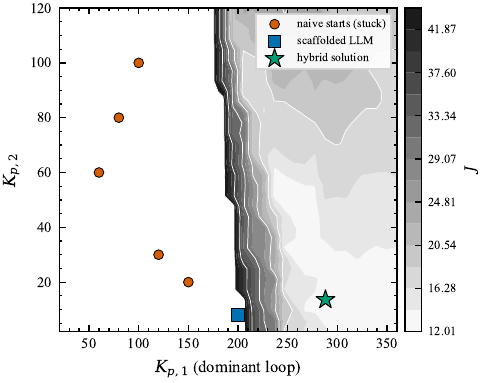}
\caption{A two-dimensional slice of the penalized cost $J$ over the proportional gains
$K_{p,1},K_{p,2}$ (integral gains fixed near the hybrid solution; lighter is lower $J$). Naive
balanced starts (circles) occupy a plateau separated by a ridge from the off-diagonal valley
that contains the scaffolded-LLM solution (square) and the refined hybrid solution (star). This
slice is centered on the asymmetric basin our method converges to; a separate, better basin
exists at different integral gains this slice does not show, reliably located by VRFT rather
than by our LLM pipeline (Sec.~\ref{sec:landscape-note}, Sec.~\ref{sec:vrft}).}
\label{fig:landscape}\end{figure}

We check that this contrast does not hinge on the specific $J<20$ threshold used to define the
``good'' region (Table~\ref{tab:main}). The ten naive-start runs span $J\in[23.4,30.2]$ and the
ten scaffolded-LLM runs span $J\in[16.6,17.1]$: a gap of $6.4$ $J$-units with no overlap, so
every threshold between $17.1$ and $23.4$ yields the identical classification (naive-starts
$0/10$, scaffolded-LLM $10/10$); $20$ is simply the midpoint of this gap, not a tuned choice.
We report Wilson $95\%$ intervals with an explicit caveat about what they can and cannot
mean here. For the scaffolded LLM and hybrid, the $10$ trials are genuine independent draws
(different decoding seeds under fixed sampling temperature), so the interval is a proper,
if small-sample, estimate of reliability under repeated use: $10/10$ ($72$--$100\%$). The
``ten naive starts,'' by contrast, are a fixed, hand-selected grid of balanced gain
vectors, not a random sample from a defined distribution of initializations; treating its
$0/10$ as an estimate of ``the true failure probability of naive initialization in general''
would overstate what a deterministic grid can support. We report its interval, $0/10$
($0$--$28\%$), as a description of \emph{this} grid's outcome, not an inferential claim about
all possible naive starts---the non-overlapping $J$ ranges in the next paragraph are the
stronger, assumption-free evidence for this specific comparison. The Latin-hypercube
baseline sits in between: its points are randomly drawn (so an interval is meaningful) but
from one design, not repeated redesigns, giving $1/10$ ($2$--$40\%$) (Table~\ref{tab:budget}).
Fig.~\ref{fig:landscape} makes the geometry explicit on a two-dimensional slice of the cost
through the proportional gains. The naive balanced starts (orange) sit on a moderate
plateau that is separated from the low-cost valley (light, right) by a steep ridge; local
descent from those starts cannot cross it, which is why $10/10$ stall. The optimum and the
LLM's proposal both lie in the off-diagonal valley ($K_{p,1}\!\gg\!K_{p,2}$)---the region a
balanced initialization never explores but that follows directly from reasoning about the
coupling.

\subsection{The hybrid: structure from the LLM, magnitudes from the optimizer}
\label{sec:hybrid}
The two failures are complementary. The optimizer cannot locate the right basin from naive
starts; the LLM locates the basin but its raw controller, while reliable ($J~16.9$), is not
the smoothest attainable. Initializing the optimizer from each of the LLM's ten
seed solutions, local refinement reaches $J~12.0\pm0.16$ (IAE~$9.1$, TV~$3.9$) in $10/10$
cases---versus $0/10$ from naive starts---approaching the $11.9$ asymmetric-basin reference. Tellingly, the refined
controller applies a small \emph{negative} integral gain on the subordinate loop, an
unintuitive coordination that decentralized tuning structurally cannot produce. Neither
component alone suffices; together they solve the problem reliably.

\subsection{Could a global optimizer replace the LLM?}
\label{sec:global}
The optimizer failures above use \emph{local} Nelder--Mead from balanced starts, so a fair
question is whether a \emph{global} optimizer would find the basin without any LLM. It would.
Differential evolution (DE) over the full gain bounds reaches the optimum region in $3/3$
independent runs ($J\!\approx\!11.9$) at a large, budget-matched run of $\sim\!540$
evaluations---close to the LLM-plus-refinement pipeline's own cost; a larger budget does not
improve on this. (As the budget sweep below shows, DE first becomes reliable, $3/3$, at a
smaller $\sim\!360$-evaluation budget, with $J~13.3$; the $\sim\!540$-evaluation run reported
here lets DE run further past that reliability threshold to approach the optimum more
closely.) A ten-start
Latin-hypercube Nelder--Mead that explicitly samples asymmetric initializations does
\emph{not} suffice ($1/10$), so the benefit is specific to a genuinely global method, not to
restarting a local one. We also test the standard sample-efficient black-box alternative,
Bayesian optimization (Gaussian-process, expected-improvement): over $10$ independent runs
at a $40$-evaluation budget it reaches the good region ($J<20$) in only $2/10$ (mean
$J~23.9$); at $150$ evaluations, $2/5$; at $250$ evaluations, $4/5$ (mean $J~18.5$)---worse
than DE at comparable or smaller budgets throughout. Bayesian optimization does not beat
differential evolution on this landscape. \emph{On this $2\times2$ plant the LLM is not
the only way to reach the optimum}---a global optimizer is an equally reliable, non-LLM
route, and (Sec.~\ref{sec:vrft}) neither is it the cheapest one.

Relative to \emph{black-box} search specifically, what distinguishes the LLM is
\emph{sample efficiency} and \emph{interpretability}. Table~\ref{tab:budget} compares
methods by evaluation budget. The scaffolded LLM returns a \emph{usable} controller
($J~16.9$) in just $18$ closed-loop evaluations; at a comparably small budget
($\sim\!40$ evaluations) DE returns $J~25.6$ and BO returns $J~23.9$---both worse than open
loop---and DE needs an order of magnitude more evaluations ($\sim\!360$) before it reliably
reaches the optimum region (BO needs more still and does not clearly converge within the
budgets we tested). The LLM also explains its choice (which loop dominates, and why),
whereas DE and BO return only numbers. We are precise about the scope of this claim: it
concerns the LLM's advantage over \emph{black-box, iterative} search specifically. It is not
an advantage over every non-LLM alternative---Sec.~\ref{sec:vrft} reports a one-shot,
non-iterative classical method that beats the full hybrid on both cost and quality---and even
within this comparison, the \emph{full hybrid} itself spends $\sim\!420$ evaluations,
comparable to DE's $\sim\!360$, without reaching the optimum more cheaply than DE.
\begin{table}[t]
\centering
\caption{Budget to a good controller (quadruple-tank, $J$). Among \emph{black-box, iterative}
methods, the LLM reaches a usable controller most cheaply. Not cheapest overall: Table~6
reports VRFT, a one-shot method beating every row here on final quality. Evals column: closed-loop
count, and (in parentheses) equivalent plant-interaction steps in thousands.}
\label{tab:budget}
\scriptsize
\begin{tabular}{lcccc}
\toprule
\textbf{Method} & \textbf{Evals (k steps)} & \textbf{$J$} & \textbf{Rel.} & \textbf{NL?} \\
\midrule
Local NM (naive)        & $158$ ($19.0$)$^{\dagger}$ & ---  & 0/10  & no \\
LHS multi-start (asym.)  & $174$ ($20.8$)$^{\dagger}$ & 15.8 & 1/10  & no \\
Diff.\ evolution         & $40$ ($4.8$)$^{\ddagger}$   & 25.6 & 0/3   & no \\
Diff.\ evolution         & $360$ ($43.2$)$^{\ddagger}$  & 13.3 & 3/3   & no \\
Bayesian opt.            & $40$ ($4.8$)   & 23.9 & 2/10  & no \\
Bayesian opt.            & $250$ ($30.0$)  & 18.5 & 4/5   & no \\
Scaffolded LLM (ours)    & $18$ ($2.2$)$^{\S}$        & 16.9 & 10/10 & yes \\
\textbf{LLM $+$ refine.\ (ours)} & $420$ ($50.4$)$^{\S}$ & \textbf{12.0} & \textbf{10/10} & yes \\
\bottomrule
\multicolumn{5}{l}{\footnotesize $^{\dagger}$measured mean \texttt{nfev} (not a cap); $^{\ddagger}$nominal DE}\\
\multicolumn{5}{l}{\footnotesize budget, not re-measured; $^{\S}$excludes $18$ separate LLM calls.}\\
\multicolumn{5}{l}{\footnotesize NL?: emits a natural-language rationale (not checked for}\\
\multicolumn{5}{l}{\footnotesize faithfulness). $1$ eval $=120$ plant steps, exact throughout.}
\end{tabular}
\end{table}

This sample-efficiency gap \emph{widens} with plant dimension, which is decisive because a
global search must cover a $2N$-dimensional gain space while the LLM's structural inference
is a single step regardless of $N$. We verify this on a synthetic $3\times3$ strongly
coupled plant, specified in full here since Table~\ref{tab:scale} alone does not make it
reproducible. The linear dynamics are $\dot y=-y+Gu$ with cyclic-dominant gain matrix
\begin{align}
G&=\begin{bmatrix}0.2&0.9&0.1\\0.1&0.2&0.9\\0.9&0.1&0.2\end{bmatrix},\\
\mathrm{RGA}(G)&=\begin{bmatrix}-.015&1.04&-.025\\-.025&-.015&1.04\\1.04&-.025&-.015\end{bmatrix},
\end{align}
so each output is dominantly driven by a \emph{different} input in a cyclic pattern
($u_1\!\to\!y_3$, $u_2\!\to\!y_1$, $u_3\!\to\!y_2$, matching the RGA's off-diagonal near-unity
entries and negative diagonal), discretized at $\Delta t{=}0.2$\,s for $K{=}120$ steps
($24$\,s), with conflicting set-points $\mathrm{SP}_1{=}(0.6,-0.4,0.5)$ switching to
$\mathrm{SP}_2{=}(-0.4,0.6,-0.3)$ at the midpoint, position-form PI per (cyclically paired)
loop with $u\in[-10,10]$, and the same penalized objective ($\lambda{=}0.75$). The reference
value $J{=}2.10$ is well-identified, not an artifact of an under-searched landscape: $6$
independent large-budget DE runs and a further $20$-point broad local multi-start all
converge to the identical point ($J{=}2.1027$ to four significant figures)---unlike the
$2\times2$ case (Sec.~\ref{sec:landscape-note}), we found no second basin here. Across
$10$ independently seeded on-premise Qwen3-14B runs, the scaffolded-only results are
$J{=}4.61\pm0.98$ (range $[3.13,6.77]$). Refining each with the same local optimizer
reaches $J{=}2.17\pm0.11$ (range $[2.10,2.43]$) in $9/10$ runs below the fixed
$J<2.42$ success threshold; the one remaining run reaches $J{=}2.434$. The mean
refinement cost is $276$ evaluations ($294$ including the $18$ scaffold evaluations per
seed), so this is a replicated $n{=}10$ dimension-scaling result rather than the earlier
two-run pilot. We did not test VRFT here (Sec.~\ref{sec:landscape-note}).
A global optimizer now needs $\sim\!2400$ evaluations
to reach the reference region---versus $\sim\!360$ on the $2\times2$ plant---whereas the scaffolded
LLM still returns a useful structural guess in $18$ evaluations (better than DE at
$\sim\!360$), and the LLM-plus-refinement hybrid reaches the near-optimal region in
about $294$ evaluations in $9/10$ runs. The hybrid's edge over global search thus grows
from negligible at $2\times2$ ($\sim\!420$ vs $\sim\!360$) to roughly $8\times$ at
$3\times3$ ($\sim\!294$ vs $\sim\!2400$).
This is a comparison against \emph{iterative black-box search} (DE) specifically; we did not
re-run VRFT at $3\times3$, so we do not know whether the one-shot data-driven route we report
in Sec.~\ref{sec:vrft} would preserve, widen, or close this margin at higher dimension---an
open question we flag rather than paper over.
\begin{table}[t]
\centering
\caption{Sample efficiency vs.\ plant dimension. Evals, with equivalent plant-interaction
steps in thousands in parentheses ($1$ eval $=120$ steps, both plants). The structural
prior's advantage over global search grows with dimension. The $3\times3$ naive-NM
denominator is $20$ (vs.\ $10$ for $2\times2$; re-run with twice the starts for a tighter
CI, same conclusion). Naive-NM: measured mean \texttt{nfev}. DE: nominal budget
(Table~\ref{tab:budget}). The LLM rows use $10$ independent decoding seeds; the hybrid
row reports the $9/10$ success rate at $J<2.42$.}
\label{tab:scale}
\scriptsize
\begin{tabular}{lcc}
\toprule
\textbf{Method} & \textbf{$2\times2$ (four-tank)} & \textbf{$3\times3$} \\
\midrule
Naive local NM (reliable)       & $0/10$, $158$ ($19.0$k)  & $7/20$, $597$ ($71.6$k) \\
Diff.\ evolution (to opt.)& $\sim$360 ($43.2$k)      & $\sim$2400 ($288.0$k) \\
Scaffolded LLM (18 evals)       & $J~16.9$       & $J~4.61\pm0.98$ \\
\textbf{LLM $+$ refine.\ (near-opt.)} & $\sim$420 ($50.4$k) & $\sim$294 ($35.3$k) \\
\bottomrule
\end{tabular}
\end{table}
Table~\ref{tab:scale} reports the DE budget at which it first becomes reliable; the full
convergence curve behind that number, now measured over $10$ independent DE runs per budget
(Table~\ref{tab:scale3curve}), makes the transition explicit and gives Wilson $95\%$
confidence intervals on each success rate rather than a single-digit fraction.
\begin{table}[t]
\centering
\caption{DE convergence on the $3\times3$ plant: success rate (Wilson $95\%$ CI) and best/mean
$J$ over $10$ independent runs per evaluation budget. Success is $J<2.42$ ($1.15\times$ the
$2.10$ reference value), fixed before the sweep; this $15\%$-margin threshold is specific to
this plant and is not the same numeric convention as the $2\times2$ tables, which use an
absolute margin (Table~\ref{tab:main}) chosen from the gap between naive and scaffolded
results---see Sec.~\ref{sec:landscape-note} for why we do not treat any single threshold as
canonical. Reliability turns on sharply between $1224$ and $2376$ evaluations.}
\label{tab:scale3curve}
\small
\begin{tabular}{lccc}
\toprule
\textbf{Evals (steps)} & \textbf{Success (95\% CI)} & \textbf{Best $J$} & \textbf{Mean $J$} \\
\midrule
$360$ ($43.2$k)  & $0/10$ ($0$--$28\%$)   & $6.72$ & $38.72$ \\
$648$ ($77.8$k)  & $0/10$ ($0$--$28\%$)   & $2.59$ & $5.71$  \\
$1224$ ($146.9$k) & $3/10$ ($11$--$60\%$)  & $2.19$ & $2.54$  \\
$2376$ ($285.1$k) & $10/10$ ($72$--$100\%$)& $2.11$ & $2.12$  \\
$4680$ ($561.6$k) & $10/10$ ($72$--$100\%$)& $2.11$ & $2.11$  \\
\bottomrule
\end{tabular}
\end{table}

\subsection{Could a direct data-driven method replace the LLM?}
\label{sec:vrft}
Every baseline so far either ignores the measured coupling (naive relay, naive local search)
or searches iteratively against many simulated closed-loop trials (DE, BO). Virtual
Reference Feedback Tuning (VRFT)~\cite{campi2002vrft} does neither: it fits a controller
from a \emph{single} batch of input--output data via one linear least-squares problem, with
no LLM and no closed-loop search in the fitting step itself. This is the most direct
classical alternative to an LLM-based structural prior, since both start from the same
premise---no first-principles model required---so a benchmark that omits it is not
answering the question it claims to answer.

\textbf{Method.} We collect one open-loop experiment: an $8$-level pseudo-random step
sequence on each pump ($48$ steps per level, long enough to approach steady state),
recording $h_1,h_2$. For each loop (the same cross pairing used throughout: $v_1\!\to\!h_2$,
$v_2\!\to\!h_1$), we choose a first-order discrete reference model
$M(z)=(1-a)/(1-az^{-1})$ with time constant $\tau$.
\begin{samepage}
The virtual reference and controller fit are defined as follows.
\[
\begin{aligned}
r_v[k]&=\frac{y[k]-a\,y[k-1]}{1-a},\\
e_v&=r_v-y,\\
K_p\,e_v[k]+K_i\sum_{j\le k}e_v[j]\Delta t&\approx u[k].
\end{aligned}
\]
\end{samepage}
where the first expression is the causal inverse of $M$. This yields four gains directly, at the cost of one experiment plus a
scan over $\tau$ (each value costs one closed-loop evaluation to check).

\textbf{Results.} Table~\ref{tab:vrft-tau} sweeps $\tau$ and shows it is not a free
parameter: quality is non-monotonic, ranging from $J~29.7$ ($\tau=15$) down to
$J~14.2$ ($\tau=200$) and back up to $J~17.8$ ($\tau=600$). At its best setting, the raw
VRFT fit ($J~14.2$) already beats naive relay-ZN ($28.6$), naive LLM ($29.7$), and is close
to the scaffolded LLM alone ($16.9$)---from one experiment and zero closed-loop search.
\begin{table}[!t]
\centering
\caption{VRFT sensitivity to the reference-model time constant $\tau$ (seconds), one
open-loop experiment, four-tank plant. Quality is non-monotonic in $\tau$; this scan is the
one design choice VRFT requires and costs $11$ evaluations, once.}
\label{tab:vrft-tau}
\begin{tabular}{cccc}
\toprule
$\tau$ & $J$ & IAE & TV \\
\midrule
15  & 29.73 & 24.40 & 7.11 \\
25  & 28.38 & 23.69 & 6.25 \\
40  & 26.37 & 23.10 & 4.36 \\
60  & 17.86 & 16.24 & 2.17 \\
90  & 15.82 & 14.87 & 1.26 \\
120 & 14.98 & 14.32 & 0.88 \\
150 & 14.45 & 13.94 & 0.67 \\
\textbf{200} & \textbf{14.23} & 13.86 & 0.48 \\
300 & 15.29 & 15.04 & 0.33 \\
400 & 16.35 & 16.15 & 0.27 \\
600 & 17.80 & 17.65 & 0.19 \\
\bottomrule
\end{tabular}
\end{table}

Refining the best-$\tau$ VRFT fit with the same local optimizer used everywhere else in this
paper does not converge to the asymmetric basin our LLM pipeline targets. Across $10$
independent open-loop experiments (different random excitation seeds, $\tau=200$ fixed),
refinement reaches $J<13$ in $10/10$ cases, mean $J=11.12\pm0.05$, range $[11.10,11.24]$
(Table~\ref{tab:vrft}): the same $10/10$ success rate as the LLM-plus-refinement hybrid but
with far tighter spread (hybrid: $\pm0.16$) and at a \emph{lower} cost. Nine of ten seeds converge to
essentially the same point, $(K_{p,1},K_{i,1},K_{p,2},K_{i,2})\approx(34,0.13,38,1.1)$---the
same near-balanced, near-zero-integral basin independently identified in
Sec.~\ref{sec:landscape-note} and, by chance once, by the inverse-gain heuristic
(Sec.~\ref{sec:heuristic}). VRFT does not merely match our best result; on this plant it is
the best method we tested, full stop, at a total cost (one experiment, an $11$-point $\tau$
scan, and a mean $258$ refinement evaluations) below the hybrid's $\sim\!420$.
\begin{table}[t]
\centering
\caption{VRFT $+$ local refinement, $10$ independent open-loop experiments. Rows 2--3
(Sec.~\ref{sec:llm-tau}) replace the $11$-point $\tau$ scan with, respectively, one LLM call
and one line of arithmetic on data already collected---the deterministic rule matches quality
and beats the LLM's reliability at zero extra cost. Row 4 is the LLM-plus-refinement hybrid
(Table~\ref{tab:main}), which all VRFT variants beat. \textbf{Evals} counts closed-loop
episode evaluations only (matching Table~\ref{tab:budget}'s convention); the one open-loop
experiment every VRFT variant also requires is a different, non-comparable resource
(raw plant steps, not episodes) and is reported separately in the footnotes, not folded into
the Evals column.}
\label{tab:vrft}
\small
\begin{tabularx}{\columnwidth}{@{}>{\raggedright\arraybackslash}Xcccc@{}}
\toprule
\textbf{Method} & \textbf{Evals} & \textbf{$J$} & \textbf{Rel.} & \textbf{Interp.} \\
\midrule
VRFT (grid-scanned $\tau$) & $\sim$270$^{\dagger}$ & 11.12$\pm$0.05 & 10/10 & no \\
LLM-guided VRFT & $\sim$250$^{\ddagger}$ & 11.19$\pm$0.17 & 8/10$^{\ddagger}$ & $\tau$ only \\
\textbf{VRFT (median-$\tau$ rule)} & $\sim$260$^{\S}$ & \textbf{11.17$\pm$0.15} & \textbf{10/10} & no \\
LLM hybrid (Table~\ref{tab:main}) & $\sim$420 & 12.0$\pm$0.16 & 10/10 & yes \\
\bottomrule
\multicolumn{5}{@{}p{\columnwidth}@{}}{\footnotesize
$^{\dagger}$Evals $=$ $11$-point $\tau$ scan $+$ mean $258$ refinement
evaluations; plus, separately, $1$ open-loop experiment ($383$ plant steps,
not an episode evaluation).
$^{\ddagger}$Evals $=$ $1$ LLM call (not an episode evaluation) $+$ mean
$248$ refinement evaluations; plus the same open-loop experiment. $2/10$
seeds exhausted the LLM's generation budget without committing to a $\tau$,
counted as failures.
$^{\S}$Evals $=$ mean refinement evaluations only (no LLM call and no extra
plant evaluation beyond the same open-loop experiment).}
\end{tabularx}
\end{table}

\textbf{What this does and does not mean.} We do not read this as ``VRFT is simply
better''---it is landscape- and method-specific, and we have not tested it beyond this
$2\times2$ plant. But it directly answers Q3: on our central pathological case, the fair
classical alternative that shares the LLM's ``no first-principles model'' premise is not
merely competitive, it wins, and we are not aware of a principled reason to expect this
plant to be unusually favorable to VRFT. What survives for the LLM is narrower and more
precise than a raw efficiency or quality claim: (i) VRFT requires choosing $\tau$, and
getting it wrong costs dearly \emph{before refinement} (a $2\times$ swing in raw $J$ across
our scan, Table~\ref{tab:vrft-tau}); Sec.~\ref{sec:llm-tau} shows this liability shrinks
substantially, but does not vanish, once the same refinement step is applied. The LLM
requires no analogous per-plant numerical hyperparameter in its scaffolding prompt.
(ii) VRFT's one-shot fit is direct numbers with no stated rationale; the LLM's proposal comes
with a natural-language account of which loop should dominate and why (Interp. column,
Table~\ref{tab:vrft}), which is auditable even if we do not verify its faithfulness
(Sec.~\ref{sec:heuristic} footnote). (iii) VRFT's excitation experiment is richer than the
single-point step tests our LLM pipeline uses to build its coupling summary; whether a
plant exists where such an experiment is unsafe or infeasible but simple step tests remain
acceptable is a real practical distinction, but we do not construct or test such a plant
here, so we state it as a hypothesis, not a finding.

\subsection{Is the LLM needed to remove VRFT's one remaining weakness?}
\label{sec:llm-tau}
Item (i) above is VRFT's only real liability: choosing $\tau$. We first re-examined
Table~\ref{tab:vrft-tau}'s $11$-point scan \emph{after} applying the same local refinement
used everywhere else in this paper, one seed, all $11$ values. Sensitivity survives
refinement only at the extreme low end---$\tau{=}15$s refines to $J{=}29.5$ (no better than
unrefined) and $\tau{=}25$s to $J{=}14.8$, both because the raw fit at these settings implies
implausibly large, aggressive gains that put the optimizer in a different, worse basin.
Across the remaining range, $\tau\in[60,600]$s---a full order of magnitude---refinement
reliably reaches the same near-optimal basin regardless of the exact value, $J\in[11.10,11.70]$.
So the liability is narrower than Table~\ref{tab:vrft-tau} alone suggests: $\tau$ only costs
dearly when it is so small that it demands closed-loop response far faster than the open-loop
plant itself moves---exactly the choice a data-grounded selection procedure should avoid.

This motivates a direct test: can the LLM propose $\tau$ from the same open-loop data VRFT
already collects, replacing the $11$-point scan with one call? For each of the $10$ open-loop
experiments behind Table~\ref{tab:vrft}, we computed, per loop and per excitation level, the
time the step response took to reach $63.2\%$ of that level's eventual change (a standard
time-constant estimate), gave these summary numbers---not raw traces---to the same on-premise
Qwen3-14B used throughout the paper together with a short account of what $\tau$ controls and
why getting it wrong is costly, and asked it to reason about and propose one shared $\tau$. It
committed to an answer in $8/10$ seeds within a fixed generation budget; the other $2$
exhausted the budget mid-reasoning without concluding, which we count as failures rather than
retrying with a larger budget until it succeeds. All $8$ proposed values fell in $[70,100]$s---
comfortably inside the range refinement rescues and far from the dangerous low extreme---despite
each one individually looking far worse than the scan optimum before refinement (raw $J$ up to
$46$, versus $14.2$ at the scanned $\tau{=}200$). After the identical refinement step, all $8$
reached the near-optimal basin: mean $J=11.19\pm0.17$ (range $[11.10,11.63]$), close to the
$10/10$, $J=11.12\pm0.05$ obtained via the full scan (Table~\ref{tab:vrft}).

Before crediting the LLM with anything, we asked the question this whole paper is organized
around one more time: is a language model actually necessary to do this, or would a trivial
rule on the same data do as well? We computed, for the same $10$ open-loop experiments, the
\emph{median} of the same per-level $63.2\%$ rise-time numbers we had handed to the LLM, and
used that single number directly as $\tau$---no reasoning, no model, no generation budget to
exhaust. This deterministic rule succeeded on $10/10$ seeds (not $8/10$) and, after the
identical refinement step, reached mean $J=11.17\pm0.15$ (range $[11.10,11.58]$). Because all
three routes (grid scan, LLM, median) refine the \emph{same} $10$ open-loop experiments, we
ran paired Wilcoxon signed-rank tests on the per-seed differences rather than treating the
means as independent. The median rule is not distinguishable from the LLM's successful seeds
($n{=}8$ paired, $p=0.69$) or from the full grid scan ($n{=}8$, $p=0.15$); the grid scan is,
in fact, paired-significantly better than the median rule ($n{=}10$, $p=0.049$), though the
effect is a mean $\Delta J=0.05$---an order of magnitude below any difference we treat as
practically meaningful elsewhere in this paper, and within the noise of the refinement
procedure itself (two seeds account for nearly all of it; the rest agree to three decimal
places). We report the significant test alongside the practically-null effect size rather
than picking whichever framing looks better.

We therefore do not credit the LLM here. A one-line statistic on data VRFT already gathers
matches the LLM's quality and beats its reliability, at no additional cost and with no failure
mode. This is the same pattern Sec.~\ref{sec:vrft} established at the level of the whole
tuning problem, now repeated one level down at the level of VRFT's single remaining design
choice: every time we gave the LLM a fair, simpler competitor to beat, something simpler won.
We report the LLM result in full above rather than deleting it, because a negative finding
about where LLM reasoning does \emph{not} help is itself the kind of evidence a benchmark
exists to produce---and because a reader attempting a similar reasoning-over-data approach
elsewhere should know a cheap baseline is the first thing to rule out, not an afterthought.

\subsection{Could a simple heuristic replace the LLM?}
\label{sec:heuristic}
If the LLM only re-derives a textbook rule, a hand-coded heuristic should match it. We test
this with a non-LLM structural prior built directly from the measured gain matrix: an
inverse-gain rule ($K_{p}\!\propto\!1/g$ on the dominant pairing) over a grid of scale and
integral-ratio settings, each refined by the same local optimizer. The answer is
\emph{landscape-dependent}, and it sharpens our boundary. On the pathological four-tank, the
heuristic prior reaches a good result in only $1/15$ configurations (best $J~11.1$, mean
$26.2$). Because the rule scales both loops by the same constant ($K_p\propto1/g$ on each
loop), its candidates are inherently near-balanced; its one success does not land on the
strongly asymmetric, negative-integral basin the LLM and hybrid target (Sec.~\ref{subsec:hybridmethod}),
but on the separate, better near-balanced basin discussed in Sec.~\ref{sec:landscape-note}
and reliably located by VRFT (Sec.~\ref{sec:vrft}); this heuristic's own single, unrefined
sweep finds it only by chance, once, in $1$ of $15$ configurations. The heuristic never
finds the asymmetric basin at all, whereas the LLM is $10/10$ within \emph{that} basin. On
the more regular $3\times3$ cyclic plant, where the coordinated structure
\emph{is} close to an inverse-gain pairing, the heuristic becomes competitive ($9/15$). This
is exactly the expected pattern: the LLM's advantage is concentrated where the optimal
coordination is \emph{non-obvious} (not recoverable by a simple rule), and fades where a rule
already captures it---the same pathological-versus-benign boundary that runs through the
paper.

A stronger classical alternative, since the full gain matrix $G$ is measured, is to decouple
explicitly rather than merely re-pair: apply the static decoupler $D=G^{-1}$ to independent
PI controllers on the natural (non-cross) pairing, so that at steady state the closed loop
sees an approximately diagonal plant, and refine from ten naive starts with the same local
optimizer used throughout. This is a substantially more informed baseline than the inverse-gain
heuristic, since $D$ uses the full off-diagonal structure of $G$, not just the dominant entries.
It is still unreliable: only $1/10$ starts reach $J<13$ and $3/10$ reach the good region
($J<20$; best $J~11.3$, mean $26.9$)---worse than the scaffolded LLM's $10/10$. (This best
value is not directly comparable to the asymmetric-basin reference in Table~\ref{tab:main}: the
decoupler adds cross-terms outside the plain per-loop PI parameterization used everywhere
else in this paper, i.e.\ a strictly richer controller class.) Static
decoupling linearizes only the measured \emph{steady-state} gain; it does not correct for the
plant's nonlinear, dynamic cross-coupling (the square-root outflow terms in
\eqref{eq:fourtank}), so the decoupled loops still interact transiently, and a fixed $D$
computed once from a step test cannot adapt to that. This result is consistent with the rest of
this section: information about the coupling helps, but only a method that reasons about
\emph{how} to use it---rather than applying a fixed linear-algebra transform or a fixed
proportionality rule---reliably finds the coordinated solution here.

\subsection{Could an enumerated structural search replace the LLM?}
\label{sec:enum}
The heuristic above tests a single, symmetric-scale rule. A sterner test is to explicitly
\emph{enumerate} the discrete choice the LLM is credited with---which loop dominates---and a
coarse grid of magnitudes and dominance ratios, then refine the best candidate with the same
local optimizer used everywhere else in this paper. We build $40$ such candidates: both
dominance patterns (loop~1 or loop~2 dominant) $\times$ dominant-loop gains
$K_p\in\{50,100,150,200,300\}$ (spanning the same bound used by the differential-evolution
baseline) $\times$ dominant-to-subordinate ratios $r\in\{5,10,20,50\}$, with
$K_i=0.05K_p$ on both loops. We test two ways of spending budget on this grid. \emph{Exhaustive}:
refine all $40$ candidates and keep the best. \emph{Screen-then-refine}: cheaply evaluate all
$40$ raw candidates ($40$ evaluations, no refinement) and refine only the single best one.

The two variants tell different stories. Exhaustive refinement reaches the optimum region
($J<13$) in only $28/40$ ($70\%$) of candidates, at a total budget of $8561$
evaluations---less reliable than the LLM's $10/10$ and roughly $20\times$ the hybrid's cost,
reinforcing the pattern from Sec.~\ref{sec:heuristic}. Screen-then-refine is the more
interesting case: it reaches $J=12.15$, matching the hybrid's quality, at a total budget of
just $198$ evaluations ($40$ screening $+\,158$ refinement)---\emph{less} than the hybrid's
$\sim\!420$. Taken at face value, this looks like a cheaper non-LLM route to the same answer.

We report it because it is real, and because the reason it works is informative. It succeeds
because the ratio grid happens to bracket the true dominant-to-subordinate ratio (the winning
candidate uses $r=20$, close to the $\sim\!20$--$30\times$ ratio the target basin actually requires);
a grid whose range did not happen to cover this scale, or a plant whose true ratio fell between
grid points, would not necessarily succeed on the first try. Building this grid already used
the measured gain matrix to bound the search (as the LLM's prompt does), but a human still had
to choose \emph{how finely} to discretize the ratio axis and confirm after the fact that the
grid bracketed the answer---the LLM performs the analogous step, inferring the right order of
magnitude for the ratio from the same coupling measurements, adaptively and without a
hand-tuned grid, and does so reliably across all $10$ seeds rather than in one favorably-designed
trial. We therefore state the sample-efficiency claim more precisely: the LLM's distinctive
value is not that no cheaper non-LLM route can ever be found post hoc, but that it needs no
manually engineered search space---it turns raw coupling measurements directly into a
correctly-scaled structural candidate, on every seed, without a designer first bracketing the
answer.

\subsection{Interpreting the learned structure}
It is worth making concrete what ``structural prior'' means here, because it ties the
empirical result back to the plant physics of \eqref{eq:fourtank} and the gain matrix $G$.
The cross-feeding of the upper tanks makes pump $v_1$ act dominantly on $h_2$ and $v_2$ on
$h_1$, so with the conflicting set-points ($h_1$ up, $h_2$ down) the two loops must
\emph{cooperate} rather than each chase its own error---the regime the negative diagonal RGA
($\lambda_{11}=-0.06$) flags as pathological for decentralized control. Both the scaffolded
LLM and the refined hybrid place the bulk of the proportional authority on one loop (e.g.\
$K_{p}\!\approx\!200$--$290$ on the dominant loop versus $\sim\!10$ on the other), exactly
the asymmetric ``one loop leads, the other trims'' structure. The optimizer's refinement
then adds a small \emph{negative} integral term on the subordinate loop: rather than
integrating its own error, that loop gently counter-acts, letting the dominant loop drive
both levels through the coupling. This is precisely the kind of coordinated, counter-intuitive
action that a relay experiment on each isolated loop can never discover and that a balanced
optimizer start is unlikely to stumble into---yet it is a short inference from the measured
coupling, which is what the LLM supplies.

\subsection{Prompt-information ablation: is the structure inferred or planted?}
\label{sec:ablation}
To test whether the LLM merely follows a hint, we first removed from the prompt only the
sentence stating that one loop should dominate, leaving the measured coupling facts and the
required-JSON output format---including its worked example,
\texttt{\{"Kp1":200,"Ki1":10,"Kp2":8,"Ki2":0.5\}}---unchanged. Performance was unchanged
(per-seed $J\in\{16.6,17.0,16.8\}$ with vs.\ without the sentence), and the LLM still proposed
an asymmetric structure on its first attempt. We originally read this as evidence the
structural insight was inferred rather than planted. It is not: the format example itself
encodes a specific, strongly asymmetric direction and magnitude ($K_{p,1}{=}200$ vs.\
$K_{p,2}{=}8$, a $25{:}1$ ratio), and removing only the verbal sentence leaves that numerical
anchor fully intact. A reviewer correctly identified this gap; we re-ran the ablation with the
example itself neutralized, replacing it with a perfectly symmetric placeholder
(\texttt{\{"Kp1":50,"Ki1":1,"Kp2":50,"Ki2":1\}}) and no verbal dominance hint, across $5$
independent seeds under the same $18$-round protocol. The result is materially different:
final structures ranged from near-balanced ($K_{p,1}{:}K_{p,2}\approx1.0${:}$1$--$1.2${:}$1$, two
of five seeds) to mildly asymmetric ($1.7${:}$1$--$4.0${:}$1$, the other three), with none
approaching the $20$--$29{:}1$ ratio the standard prompt reliably produces
(mean $J=13.86\pm1.69$, versus the standard prompt's $16.9\pm0.2$). We therefore withdraw
the earlier claim that the LLM ``reliably'' finds \emph{this specific} structure through
reasoning alone: a meaningful part of that reliability was the format example acting as a
soft anchor, not measured coupling data. What survives, and is still genuine, is that
coupling-conditioned reasoning carries real signal even under a fully neutral prompt---mean
$J=13.86$ is far better than naive LLM tuning ($29.7$) or naive relay ($28.6$), and two of
five neutral-prompt seeds independently reached the better, near-balanced basin
(Sec.~\ref{sec:landscape-note}) that the standard, anchored prompt never reaches in $10/10$
runs. A separate prompt variant locates what \emph{is} load-bearing on top of this: replacing
the prose coupling description with only the raw gain matrix (no ``$v_1$ raises $h_2$
strongly'' phrasing) degrades performance to $J~18.6$--$27.1$ with high variance, confirming
that the coupling summary itself, not just the format example, carries real information. We
do not claim to isolate ``reasoning'' from the model's internalized control knowledge---an
open-weight LLM may well draw on learned heuristics such as RGA-based pairing---and we no
longer claim the specific asymmetric structure reported throughout this paper is a clean,
reasoning-only artifact; it is a property of the full scaffolding, prompt format included,
which is what we release and what a practitioner would actually deploy.

\subsection{Generalization across open models}
\begin{figure}[t]\centering\includegraphics[width=0.92\columnwidth]{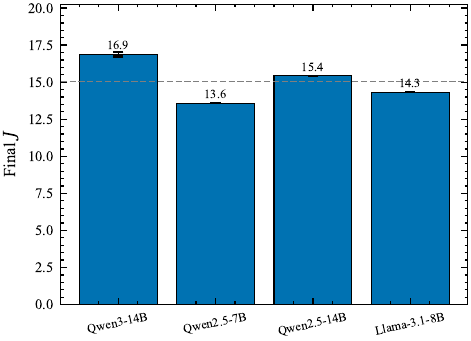}
\caption{Generalization: four open models all reach the optimum region under the same
scaffolding (penalized cost $J$).}\label{fig:multi}\end{figure}

The behaviour is not specific to one model. With identical scaffolding, four open models
spanning two families and three sizes all reach the coordinated structure
(Fig.~\ref{fig:multi}): Qwen3-14B ($J~16.9$), Qwen2.5-7B ($13.6$), Qwen2.5-14B ($15.4$) and
Llama-3.1-8B ($14.3$)---all far below naive LLM tuning ($29.7$) and open loop ($22.7$).
Notably Llama-3.1-8B discovers the smooth subordinate-loop-dominant structure directly
(TV~$1.6$), while the others find the loop-1-dominant basin; either way the optimizer refines
each to the $J~12$ region. Table~\ref{tab:models} reports the decomposition. These per-model
values are single runs (the Qwen3-14B entry is the ten-seed mean of Table~\ref{tab:main}), so
the modest ordering differences---e.g.\ Qwen2.5-7B below Qwen3-14B---fall within the
seed-to-seed spread and should not be over-interpreted; the robust statement is that every
model reaches the coordinated region. The broad coordinated-region benefit is thus a property
of the scaffolding, not of a single model; convergence to one exact structure remains
prompt-sensitive.
\begin{table}[t]
\centering
\caption{Scaffolded tuning across four open models (penalized cost $J=\mathrm{IAE}+0.75\,\mathrm{TV}$).
All reach the coordinated region; none collapses to the naive-LLM failure ($J~29.7$).}
\label{tab:models}
\begin{tabular}{lcccc}
\toprule
\textbf{Model} & \textbf{Family} & \textbf{$J$} & \textbf{IAE} & \textbf{TV} \\
\midrule
Qwen3-14B    & Qwen3   & 16.9 & 11.6 & 7.0 \\
Qwen2.5-7B   & Qwen2.5 & 13.6 & 9.5  & 5.4 \\
Qwen2.5-14B  & Qwen2.5 & 15.4 & 10.7 & 6.3 \\
Llama-3.1-8B & Llama 3 & 14.3 & 13.1 & 1.6 \\
\bottomrule
\end{tabular}
\end{table}

\subsection{Robustness to the penalty weight}
\label{sec:lambda}
The structural-prior advantage is not an artifact of the particular weight $\lambda=0.75$.
Sweeping $\lambda\in\{0.2,0.3,0.5,0.75,1.0\}$ and, for each, refining from naive balanced
starts versus from LLM-style structural priors, the qualitative picture is invariant
(Table~\ref{tab:lambda}): the naive-start optimizer reaches the optimum region in at most
$2/10$ runs at every weight, while refinement from the LLM's structural prior reaches it in
$5/5$ runs throughout. Smaller $\lambda$ keeps the high-gain (rougher) solution; from
$\lambda\!\gtrsim\!0.5$ the best basin moves to the smooth, all-positive-gain regime we adopt.
Across this range the prior's reliability advantage is stable, so $\lambda=0.75$ is a
reasonable operating point rather than a tuned sweet spot.
\begin{table}[t]
\centering
\caption{Robustness to the control-effort weight $\lambda$. ``Reliable'' is the fraction of
optimizer runs reaching the reference region; the LLM structural prior is reliable ($5/5$) at
every $\lambda$, while the naive-start optimizer reaches at most $2/10$. \textbf{Best $J$} is
the best value found in the asymmetric basin the LLM targets, not a global optimum (a
better, near-balanced basin exists at every $\lambda$; Sec.~\ref{sec:landscape-note},
Sec.~\ref{sec:vrft}).}
\label{tab:lambda}
\begin{tabular}{ccccc}
\toprule
$\lambda$ & Best $J^{\dagger}$ & (IAE/TV) & Naive starts & LLM prior \\
\midrule
0.20 & 9.9  & (9.1/3.8) & 2/10 & 5/5 \\
0.30 & 10.3 & (9.1/3.9) & 2/10 & 5/5 \\
0.50 & 11.0 & (9.1/3.8) & 0/10 & 5/5 \\
0.75 & 11.9 & (9.1/3.8) & 0/10 & 5/5 \\
1.00 & 11.9 & (10.4/1.5) & 0/10 & 5/5 \\
\bottomrule
\multicolumn{5}{l}{\small $^{\dagger}$asymmetric-basin best, not a proven global optimum.}
\end{tabular}
\end{table}

\subsection{A two-signal diagnostic for when a structural prior is worth it}
\label{sec:boundary}
Every result in this paper that favors a structural prior (LLM or VRFT) over naive local
search shares one property: the tuning landscape is \emph{pathological}, in the specific
sense that the coordinated solution is not the one naive, balanced initialization finds. A
practitioner does not need to run a full benchmark to know whether their own plant is in this
regime. The first, and genuinely free, signal is the relative gain array (RGA): computable
from a single steady-state gain measurement, before running any optimization at all. A
diagonal entry $\lambda_{ii}$ far from $1$ (including negative) signals that the natural
decentralized pairing is itself unstable or non-obvious. The second signal,
start-sensitivity---the fraction of naive, balanced initializations from which \emph{local}
optimization reaches the optimum region, holding the evaluation budget fixed across
plants---is \emph{not} free in the same sense: computing it means actually running local
optimization from several starts, which is itself a cheap tuning attempt, not a
pre-tuning measurement, and it also presupposes knowing what the optimum region looks like.
We report it as a confirmatory second signal, not as a zero-cost predictor: across the four
plants below it tracks the RGA ordering closely, but a practitioner who wants a signal
available before running anything at all should rely on RGA alone, which already reproduces
the same ordering on its own (Table~\ref{tab:diagnostic}).

We have this pair of numbers for four structurally different systems spanning a full range
from trivial to severely pathological. The single-loop CSTR (Sec.~\ref{subsec:cstr}) has no
interaction to speak of ($\lambda_{11}=1$ by construction for a SISO loop); relay-feedback
tuning reaches within $4\%$ of the numerically-found best value and the LLM has nothing to add
(it is $50\%$ worse). The quadruple-tank ($\lambda_{11}=-0.06$) is our central pathological
case: local optimization from naive starts fails in $10/10$ trials, and a structural prior
(LLM or VRFT) is necessary to reach a good basin at all. The series-reactor pair
($\lambda_{11}=1.88$) is genuinely interacting but benign: naive-start local optimization
converges in $7/7$ trials, and the LLM holds no advantage. The $3\times3$ cyclic plant
(Sec.~\ref{sec:global}, all three diagonal RGA entries $\approx-0.015$) sits between the
quadruple-tank and the benign cases on both diagnostics: naive-start local optimization
succeeds in $7/20$ trials (Table~\ref{tab:scale})---worse than the benign plants, better than
the quadruple-tank's $0/10$---and a structural prior remains valuable (the LLM-plus-refinement
hybrid reaches the near-optimal region roughly $8\times$ more cheaply than differential
evolution; this is an LLM-versus-DE comparison only, Sec.~\ref{sec:global}), though less
starkly necessary than on the quadruple-tank.
\begin{table}[t]
\centering
\caption{The two-signal diagnostic across four structurally distinct plants, ordered by
$|\lambda_{ii}-1|$ (distance of the RGA diagonal from the ideal-pairing value). Both signals
and the empirical need for a structural prior move together, monotonically, across all four.}
\label{tab:diagnostic}
\scriptsize
\begin{tabular}{lccc}
\toprule
\textbf{Plant} & \textbf{RGA} $\lambda_{ii}$ & \textbf{Naive-start} & \textbf{Prior} \\
 & & \textbf{success} & \textbf{helps?} \\
\midrule
CSTR ($1{\times}1$)      & $1.00$      & reliable$^{\dagger}$ & no \\
Series reactor ($2{\times}2$) & $1.88$ & $7/7$                & no \\
$3{\times}3$ cyclic       & $\approx-0.015$ & $7/20$           & yes$^{\ddagger}$ \\
Quadruple-tank ($2{\times}2$) & $-0.06$ & $0/10$               & yes$^{\S}$ \\
\bottomrule
\multicolumn{4}{l}{\footnotesize $^{\dagger}$relay-ZN, SISO, no start-sensitivity to speak of.}\\
\multicolumn{4}{l}{\footnotesize $^{\ddagger}$structural prior improves efficiency, not necessity.}\\
\multicolumn{4}{l}{\footnotesize $^{\S}$structural prior is necessary, not just cheaper.}
\end{tabular}
\end{table}
This is four plants, not a designed suite that sweeps RGA and start-sensitivity independently
to establish false-positive and false-negative rates---we do not claim the pattern is
exhaustively validated, only that it holds, in the same direction, on every structurally
different system we happened to test, including one ($3\times3$) added specifically to probe
the space between our two original endpoints. A practitioner can compute both numbers from a
single round of step tests, before running any tuning method at all, and use Table~\ref{tab:diagnostic}'s
ordering as a prior on whether the investment in a structural prior---LLM, VRFT, or otherwise---is
likely to pay off on their own plant.

% ============================================================
\section{Discussion and Limitations}
\textbf{What the LLM actually contributes---and what it does not.} Across the coupled-MIMO
experiments, the full scaffolded pipeline (coupling summary, objective framing, and required
JSON format together) reliably produces gains in a good asymmetric basin, across seeds and
model families, with a stated rationale attached. We initially attributed this reliability to
reasoning over the coupling data alone; Sec.~\ref{sec:ablation} shows that attribution was too
generous---the JSON format example itself measurably anchors the specific structure found, and
without it the same pipeline is markedly less consistent, though still well above naive
baselines. The honest claim is narrower than ``the LLM infers which loop must dominate'':
coupling-conditioned reasoning carries real, reproducible signal, but the crisp, highly
reliable convergence to one specific structure is a property of the prompt as a whole, format
example included, not of reasoning in isolation. It does \emph{not} produce the smoothest
attainable controller either way: the asymmetric
basin it reliably finds ($J\!\approx\!11.9$) is not the best one on this plant. A second,
near-balanced, near-zero-integral basin exists at $J\!\approx\!11.1$
(IAE~$10.1$, TV~$1.3$ vs.\ the asymmetric basin's IAE~$9.1$, TV~$3.8$)---smoother actuation
traded for slightly worse tracking---and, as Sec.~\ref{sec:vrft} shows in full, Virtual
Reference Feedback Tuning finds it \emph{reliably} ($10/10$, tighter than the LLM's own
spread) from a single open-loop experiment. We located this basin while auditing our own
asymmetric-basin reference claim (three independent search procedures converge to it to four
significant figures, confirming it is genuine, not numerical noise) and initially
mischaracterized it in an earlier draft as belonging to the asymmetric basin instead; an
earlier, narrower version of this section speculated about hand-crafted prompt hints to find
it, which we removed once VRFT gave a real, non-circular answer instead. Uniform random search
essentially never finds this basin ($0/30$ starts) and DE rarely does ($1/8$ runs); the LLM's
proposed priors never do either. So the honest summary is: the LLM reliably finds \emph{a}
good, previously-unreachable-by-naive-search basin, but not \emph{the} best one, and a
classical data-driven method finds a better one more reliably and, on this plant, more
cheaply. \emph{Local} optimization alone---excellent at refinement but blind to any good
basin---fails from balanced starts regardless of which basin is targeted; a global optimizer
(Sec.~\ref{sec:global}) escapes that trap but lands on the same asymmetric basin as the LLM,
not the better one, at far higher cost. A hand-designed grid enumeration over dominance
patterns can also reach the asymmetric basin (Sec.~\ref{sec:enum}), but only once a designer
has already bracketed the right magnitude and ratio.\label{sec:landscape-note}

\textbf{Why a control-effort penalty matters here.} Optimizing tracking error alone rewards
high-gain, near-saturating ``bang--bang'' controllers that no plant operator would deploy.
Scoring controllers by $J=\mathrm{IAE}+\lambda\,\mathrm{TV}(u)$ both reflects realistic
deployment concerns (valve wear, actuator stress) and, we found, makes the optimum
genuinely smoother. Notably it also sharpens the central result: the naive-start optimizer, which already misses
the optimum region in $7$ of $10$ runs under a pure-IAE objective, fails in all $10$ under
the penalized cost, because the smooth optimum sits in an even narrower, more
counter-intuitive region---exactly where a structural prior is most valuable.

\textbf{Why the direct route keeps winning.} A pattern runs through every comparison in this
paper: VRFT beats the LLM hybrid on the central task, a one-line median beats the LLM at
choosing VRFT's $\tau$, and the LLM's own reliability turned out to depend partly on a
numerical anchor in its prompt rather than reasoning alone. A simple reading covers all three:
VRFT and the median rule are direct, close-to-closed-form maps from measured data to the
quantity being estimated, while the LLM's route runs through a natural-language
interpretation of the same data before it ever produces a number. Where the needed structure
can be expressed as a closed-form or near-closed-form statistic of the measurements---a
least-squares fit, a median rise time---the more direct route has less room to drift, and
wins. This does not predict the LLM is never useful; it predicts \emph{where} to expect it to
struggle against a fair alternative, and every test in this paper is consistent with that
prediction.

\textbf{When to reach for an LLM---and when to reach for VRFT first.} Given
Sec.~\ref{sec:vrft} and Sec.~\ref{sec:llm-tau}, the honest practical recommendation is: try
VRFT (or another direct data-driven method) before an LLM whenever a sufficiently rich
open-loop or arbitrary input--output experiment is available, since on our central case it is
cheaper, more reliable, and reaches a better solution---and even VRFT's one remaining design
choice is better resolved by a one-line statistic on the same data than by an LLM call. What
the LLM keeps, on the evidence in this paper, is narrower than a design-choice-removal role: an
auditable, natural-language rationale for the chosen structure, where that rationale is a
deployment requirement and not merely a convenience---and even that claim is now qualified by
Sec.~\ref{sec:ablation}'s finding that the specific structure a rationale accompanies is itself
sensitive to prompt formatting. Beyond the LLM-vs-VRFT question, the value of \emph{any}
structural prior (LLM or classical) over local numerical search is still \emph{landscape-specific}:
Sec.~\ref{sec:boundary} formalizes this as a two-signal diagnostic---RGA and optimizer
start-sensitivity---that a practitioner can compute \emph{before} choosing a tuning method at
all, on any new plant.

\textbf{Practical deployment.} Because the method is designed for the on-premise setting,
its operational footprint is modest and compatible with plant IT constraints. The largest
model we use (14B parameters) runs in \texttt{bfloat16} on a single 48~GB GPU, so a tuning
campaign needs one workstation and never transmits process data off-site---directly
satisfying the data-governance rules that rule out cloud APIs. No fine-tuning is required:
the same open checkpoint is used out of the box, with capability supplied entirely by the
scaffolding prompt, which keeps the deployment auditable (the prompt and the gain proposals
are human-readable and loggable). A tuning run is a bounded, offline procedure---a fixed
number of closed-loop simulations interleaved with short LLM calls---rather than an online
controller, so its latency does not enter the control loop; the LLM is consulted at
\emph{design} time, and the artifact shipped to the plant is an ordinary PI controller.
These properties matter as much as the accuracy numbers for whether such a method can
actually be adopted in a regulated industrial environment.

\textbf{Limitations.} Our evidence spans $2\times2$ and $3\times3$ coupled plants and a PI
control structure under deterministic simulation. We do not go beyond $3\times3$, nor do we
provide formal stability margins or study hard input/state constraints or full
hardware-in-the-loop effects (see~\cite{gu2022safe} for a survey of safe learning-based
control addressing exactly these concerns), and the
LLM-as-prior idea may interact with these in ways our benchmark cannot reveal. We do test
robustness to two of these factors: performance is stable to penalty weight
(Sec.~\ref{sec:lambda}) and the hybrid degrades only gracefully when the LLM's proposed prior
is itself imprecise. We perturb each of the ten real scaffolded-LLM gain vectors with
independent multiplicative Gaussian noise, $g_i'=g_i(1+\epsilon_i)$, $\epsilon_i\sim
\mathcal{N}(0,\sigma^2)$, at $\sigma\in\{10\%,25\%,50\%\}$ relative error, then refine each
corrupted prior as usual; hybrid refinement still reaches the optimum region in $9/10$,
$7/10$, and $7/10$ of ten runs, respectively, a gradual decline rather than a cliff. This
tests robustness to an imprecise structural prior, not to noise in the measured coupling
data itself, which we leave to future work. We also did not test VRFT (Sec.~\ref{sec:vrft})
beyond the $2\times2$ plant, so whether it, rather than the LLM, retains the sample-efficiency
edge over global search at $3\times3$ and beyond is open. On-premise open
models are improving rapidly, so the competence floor we report is likely conservative.

% ============================================================
\section{Conclusion}
We asked whether on-premise open-weight LLMs are useful for controller tuning, and answered
with a deliberately honest benchmark rather than a favorable demo---one that tests the
classical alternative most likely to make the LLM unnecessary, and reports it losing when it
does. On a well-posed single loop the LLM is competent but beaten by a classical relay tuner.
On a strongly coupled quadruple-tank with conflicting set-points, naive relay tuning and naive
LLM tuning both fail to beat open loop, and local optimization from balanced starts fails in
all $10$ runs tested ($0/10$); a scaffolded LLM
reliably ($10/10$) converges on a counter-intuitive asymmetric structure that, refined by a
local optimizer, reaches $J~12.0\pm0.16$, and this holds across four open models spanning
two model families---though Sec.~\ref{sec:ablation} shows this specific reliability is a
property of the full prompt, format example included, not of reasoning over coupling data in
isolation. This is where a favorable-demo paper would stop. We did not: Virtual Reference
Feedback Tuning~\cite{campi2002vrft}, a decades-old direct data-driven method that shares the
LLM's ``no first-principles model'' premise, fits a controller from one open-loop experiment
and---refined the same way---matches the LLM-based hybrid's $10/10$ success rate with roughly
$3\times$ tighter spread and
reaches a \emph{better} solution ($J~11.1\pm0.05$), at comparable or
lower total cost. Neither a global optimizer nor Bayesian optimization closes this gap. We
report this because a benchmark that is only willing to publish favorable comparisons is not
a benchmark, and because the finding is itself useful: for this class of problem, practitioners
should try direct data-driven tuning before reaching for an LLM. VRFT's own weakness is
choosing its one design parameter $\tau$: sensitivity survives refinement only in a narrow
low-$\tau$ region. We tested whether the LLM could remove this weakness by reasoning over the
same open-loop data VRFT already collects, and it can, landing safely outside the dangerous
region in $8/10$ seeds---but before crediting it, we tested the obvious cheaper alternative: the
median of the same rise-time numbers, used directly with no model and no reasoning, succeeds in
$10/10$ seeds at matching quality. We credit the arithmetic, not the LLM, and report the LLM
result anyway, because a benchmark willing to publish only its subject's wins is not a
benchmark: the same pattern that decided the central comparison of this paper---something
simpler beating the LLM when we bothered to test it---repeats one level down, at the level of
VRFT's own remaining design choice. The benefit over naive local search, LLM or classical, is
further confined to plants whose coordinated solution is genuinely pathological, and we
contribute a concrete, a priori way to tell which regime a new plant is in before investing in
either tuning route: the RGA diagonal, computable from step-test data before running anything,
which on its own reproduces the same ordering as naive-start optimizer reliability
(Sec.~\ref{sec:boundary}) across every one of the four structurally different plants we
tested, from a trivial single loop to our most pathological case, with start-sensitivity
available as an inexpensive but non-free confirmatory check. The practical
message is not that on-premise LLMs are the tool of choice for MIMO controller tuning---our
own evidence says otherwise on the case we studied hardest---but that a rigorous benchmark
willing to let its subject lose is more useful to practitioners than one that is not, and that
the field should default to comparing LLM-based control methods against direct data-driven
tuning, not only against black-box search. Future work should test whether VRFT's advantage
persists at higher dimension, whether the median-$\tau$ rule remains sufficient there too or
whether a harder problem finally makes reasoning necessary, whether the LLM's scaffolding can
be extended to discover the better basin we found rather than only the good one it currently
finds, and whether either method can be made to carry explicit stability and robustness
guarantees. We release all code, prompts, the VRFT, Bayesian-optimization, and $\tau$-selection
(LLM and deterministic) scripts, and the per-run ledger to support replication and extension.

\section*{Reproducibility}
Core method and test-bed code, with the scaffolding prompts, are available at
\url{https://github.com/cheer932041235/llm-onprem-control}. Table~\ref{tab:repro} lists the key
settings.
\begin{table*}[t]
\centering
\caption{Reproducibility settings.}
\label{tab:repro}
\small
\begin{tabularx}{\textwidth}{@{}>{\raggedright\arraybackslash}p{0.16\textwidth}
  >{\raggedright\arraybackslash}X@{}}
\toprule
\textbf{Item} & \textbf{Value} \\
\midrule
Simulators        & PC-Gym (quadruple-tank, CSTR); custom $3\times3$ \\
Episode           & $N{=}120$ steps, $400$\,s; conflicting set-points \\
Controller        & position-form PI, cross pairing; $v\in[0.1,10]$; \\
                  & clamping anti-windup (integral held when saturated) \\
Objective         & $J=\mathrm{IAE}+\lambda\,\mathrm{TV}(u)$, $\lambda{=}0.75$ \\
LLM               & Qwen3-14B (+3 open models), \texttt{bf16}, $T{=}0.9$, $p{=}0.95$ \\
LLM tuning        & 18 rounds; 10 seeds (scaffolded, hybrid) \\
Optimizer         & Nelder--Mead (\texttt{scipy}), \texttt{xatol}${=}0.3$--$0.5$, \\
                  & \texttt{fatol}${=}10^{-3}$, \texttt{maxiter}${=}200$--$300$ \\
DE hyperparams    & \texttt{scipy} defaults (mutation $\in[0.5,1]$, \\
                  & recombination${=}0.7$), \texttt{popsize}${=}12$--$20$ \\
Gain bounds (DE)  & $K_p\in[0,400]$, $K_i\in[-10,30]$ \\
LLM output clip   & raw gains clipped to $[-50,1500]$ before simulation \\
                  & (safety net, never approached in practice$^{\P}$); \\
                  & unparseable JSON or non-finite $J$ skipped and retried \\
Seed scope        & seeds index independent LLM sampling / optimizer \\
                  & starts; simulator is deterministic (no process noise) \\
VRFT              & 1st-order ref. model, $\tau{=}200$s (scanned $15$--$600$); \\
                  & 8-level open-loop excitation, $48$ steps/level, $10$ seeds \\
LLM-guided $\tau$ & same Qwen3-14B, per-loop per-level $63.2\%$ rise-time \\
                  & summary as input; 2500-token generation budget, \\
                  & $3$ resample attempts/seed; $8/10$ seeds committed to $\tau$ \\
Median-$\tau$ rule & $\tau=$ median of the same per-loop per-level rise-time \\
                  & numbers handed to the LLM; no model, $10/10$ seeds \\
Bayesian opt.     & \texttt{scikit-optimize} \texttt{gp\_minimize}, EI acquisition, \\
                  & same bounds as DE, $n\_calls\in\{40,150,250\}$ \\
Hardware          & 1$\times$ RTX~6000 Ada (48~GB), on-premise \\
\bottomrule
\multicolumn{2}{@{}p{\textwidth}@{}}{\footnotesize $^{\P}$Every LLM-proposed gain across all $10$
scaffolded-LLM seeds falls in $K_p\in[6.5,210]$, $K_i\in[0.4,10]$; every refined hybrid
solution falls in $K_p\in[9,293]$, $K_i\in[-4.9,10]$---both well inside the DE/BO search box
($K_p\in[0,400]$, $K_i\in[-10,30]$). The LLM is never advantaged by a wider effective search
domain than the black-box baselines in practice, even though its nominal output clip is
wider.}
\end{tabularx}
\end{table*}
\FloatBarrier

\section*{Funding}
This work received no external funding.

\section*{Acknowledgment}
During the preparation of this work, the authors used Anthropic Claude to
assist with language editing and readability, and OpenAI GPT Image to produce
an initial visual draft of the conceptual illustration in Fig.~\ref{fig:concept}.
The authors substantially revised and verified the resulting material,
including every technical statement, numerical value, and figure label. No
experimental result or reference was generated by these tools. The authors
take full responsibility for the content of the publication.

\end{document}